\documentclass{article}
\usepackage[T1]{fontenc}
\usepackage{geometry}
\usepackage{graphicx}
\usepackage{xcolor}
\usepackage[font=small]{caption}
\usepackage{amsmath,amssymb,latexsym}
\usepackage{marvosym}
\usepackage{url}
\usepackage{lipsum}
\usepackage{booktabs}
\usepackage{amssymb}
\usepackage{hyperref}
\usepackage{bm}
\usepackage{float}
\usepackage[english]{babel}
\usepackage{hyperref}
\usepackage{subcaption}
\usepackage{subfloat}
\usepackage{epsf}
\usepackage{float}
\usepackage{pifont}
\usepackage{wrapfig}
\usepackage{multicol}
\usepackage{enumitem}
\usepackage{xcolor}
\usepackage[utf8]{inputenc}
\usepackage{framed}
\usepackage{textcomp}
\usepackage{braket}
\usepackage[labelfont=bf]{caption}
\usepackage{algpseudocode}
\usepackage{algorithm2e}
\usepackage[normalem]{ulem}
\usepackage[numbers,sort&compress]{natbib}
\usepackage{mathrsfs}
\usepackage{authblk}
\graphicspath{{}}
\usepackage{color, colortbl}
\usepackage{chemformula}
\usepackage{xcolor}
\usepackage{authblk}

\author[1, 2]{\normalsize Kamila Zdybał}
\author[1,2,3]{\normalsize Giuseppe D'Alessio}
\author[1, 2]{\normalsize Gianmarco Aversano}
\author[1, 2]{\normalsize Mohammad Rafi Malik}
\author[1, 2]{\normalsize Axel Coussement}
\author[4]{\normalsize James C. Sutherland}
\author[1, 2]{\normalsize Alessandro Parente}

\affil[1]{Université Libre de Bruxelles, École polytechnique de Bruxelles, Aero-Thermo-Mechanics Laboratory, Brussels, Belgium}
\affil[2]{Université Libre de Bruxelles and Vrije Universiteit Brussel, Combustion and Robust Optimization Group (BURN), Brussels, Belgium}
\affil[3]{CRECK Modeling Lab, Department of Chemistry, Materials and Chemical Engineering, Politecnico di Milano, Milan, Italy}
\affil[4]{Department of Chemical Engineering, University of Utah, Salt Lake City, Utah, USA}

\title{Advancing Reacting Flow Simulations \\ with Data-Driven Models}

\date{}

\usepackage{lineno}

\begin{document}

\maketitle

The use of machine learning algorithms to predict behaviors of complex systems is booming. However, the key to an effective use of machine learning tools in multi-physics problems, including combustion, is to couple them to physical and computer models. The performance of these tools is enhanced if all the prior knowledge and the physical constraints are embodied. In other words, the scientific method must be adapted to bring machine learning into the picture, and make the best use of the massive amount of data we have produced, thanks to the advances in numerical computing. The present chapter reviews some of the open opportunities for the application of data-driven reduced-order modeling of combustion systems. Examples of feature extraction in turbulent combustion data, empirical low-dimensional manifold (ELDM) identification, classification, regression, and reduced-order modeling are provided.
\vspace{6mm}

\vspace{6mm}

\section{Introduction}\label{section:intro}

The simulation of turbulent combustion is a very challenging task for a number of aspects beyond turbulence. Combustion is intrinsically multi-scale and multi-physics. It is characterized by a variety of scales inherently coupled in space and time through thermo-chemical and fluid dynamic interactions \cite{Pope_small_scales}. Typical chemical mechanisms describing the evolution of fuels consist of hundreds of species involved in thousands of reactions, spanning 12 orders of magnitude of temporal scales \cite{PolimiC1C3HTNOX}. The interaction of these scales with the fluid dynamic ones defines the nature of the combustion regime as well as the limiting process in determining the overall fuel oxidation rate \cite{kuo2012fundamentals}. When the characteristic chemical scales are much smaller than the fluid dynamic ones, the combustion problem becomes a mixing one (i.e., \textit{mixed is burnt} \cite{Magnussen_1981}): combustion and chemistry are decoupled, and the problem is highly simplified. Likewise, for chemical timescales much larger than the fluid dynamic ones, the system can be described taking into account chemistry only, neglecting the role of fluid dynamics altogether.

The intensity of interactions between turbulent mixing and chemistry is measured using the Damköhler number, defined as the ratio between the characteristic mixing, $\tau_m$, and chemical, $\tau_c$, timescales:
\begin{equation}\label{eq:Da}
Da=\frac{\tau_m}{\tau_c} .
\end{equation}
In terms of the Damköhler number, $Da \gg 1$ indicates a mixing-controlled, fast chemistry process. On the other hand, $Da \ll 1$ denotes a chemistry-controlled, slow chemistry process. Most practical combustion systems operate at conditions characterized by a non-negligible overlap between flow  and chemical scales. This is particularly true for novel combustion technologies, where the use of diluted conditions and the enhanced mixing leads to a $Da$ distribution close to unity. This grants some control on the combustion process, thanks to the increase of the characteristic chemical scales compared to the mixing ones. In particular, the operating conditions (temperature and compositions) can be adjusted in such a way that the emissions are kept below the required values \cite{Cavaliere}, \cite{Wunning}, \cite{parente2011investigation}. The condition $Da \approx 1$ is generally referred to as finite-rate chemistry, to indicate that combustion is not infinitely fast but of finite speed.

Modeling finite-rate combustion regimes is very challenging because both fluid mechanics and chemistry effects must be accounted for accurately. In particular, chemistry cannot be described using simplified global mechanisms, and this results in a significant computational burden of combustion simulations. The resolution of turbulent combustion problems requires the resolution of hundreds of transport equations for (tightly coupled) chemical species on top of the conservation equations for mass, momentum, and energy. Beside the high dimensionality of the problem \cite{lu2009toward}, the transport equations of reacting scalars require closure models, when the reacting structures are not fully resolved on the numerical grid.

The challenges associated with turbulent combustion modeling make the use of machine learning very attractive. While turbulent combustion models are spread across combustion industries, their current predictive capabilities fall well short of what would be needed in decision making for new design and regulation \cite{Pope_small_scales}. High-fidelity, direct numerical simulations (DNS) of combustion systems are still limited to isolated aspects of a turbulent combustion process and simple \textit{building blocks}. Still, these high-fidelity simulations are rich in information that could help decode the complexity of turbulence-chemistry interactions and guide the development of filtered and lower-fidelity modeling approaches for faster evaluations.

The objective of the present chapter is to demonstrate the potential of data-driven modeling in the context of combustion simulations. In particular, we present:
\begin{itemize}
        \item
        The application of principal component analysis (PCA) and other linear and nonlinear techniques to identify low-dimensional manifolds in high-fidelity combustion data-sets, and to reveal the key features of complex non-equilibrium phenomena. Different techniques are compared to PCA, including non-negative matrix factorization (NMF), autoencoders, and local PCA in \S\ref{section:feature}.
        \item
        The development of reduced-order models (ROMs), to be used in conjunction with, or to replace high-fidelity simulation tools, to reduce the burden associated with the large number of species in detailed chemical mechanisms. First, the use of transport models based on PCA is presented in \S\ref{section:pc_transport}. Finally, the application of the data-driven adaptive-chemistry approach based on the combination of classification and chemical mechanism reduction is discussed in \S\ref{section:sparc}.
\end{itemize}

\section{Combustion Data Sets}\label{section:data-sets}

The structure of combustion data sets differs from the one seen in pure fluid mechanics applications (presented in Chapter 4). The data set encountered in multi-component reactive flows is stored in the form of a matrix $\mathbf{X} \in \mathbb{R}^{N \times Q}$. Each column of $\mathbf{X}$ is tied to one of the $Q$ thermo-chemical state-space variables: temperature $T$, pressure $p$, and $N_s-1$ chemical species\footnote{Since mass (or mole) fractions sum up to unity for every observation, out of $N_s$ species only $N_s-1$ are independent.} mass (or mole) fractions, denoted $Y_i$ for the $i$th species. For open flames and atmospheric burners the pressure variable can be omitted. Each of the $N$ rows of $\mathbf{X}$ contains observations of all $Q$ variables at a particular point in the physical space and/or time (and sometimes, a point in the space of other independent parameters, as will be briefly discussed later). This structure of the data matrix is presented below:
\begin{gather}\label{eq:data-set}
\mathbf{X} = 
\begin{bmatrix}
\vdots & \vdots & \vdots & \vdots & & \vdots  \\
T & p & Y_1 & Y_2 & \dots & Y_{N_s - 1}  \\
\vdots & \vdots & \vdots & \vdots & & \vdots  \\
\end{bmatrix} .
\end{gather}
For such data set, $Q = N_s+1$. We will denote the $i$th row (observation) in $\mathbf{X}$ as $\mathbf{x}_i \in \mathbb{R}^{Q}$ and the $j$th column (variable) in $\mathbf{X}$ as $\mathbf{X}_j \in \mathbb{R}^{N}$. When the data set is only resolved in space and not resolved in time, $N$ represents the number of points on a spatial grid, and $\mathbf{X}$ can be thought of as a data \textit{snapshot} (a notion much like the one discussed in Chapter 4). Typically, we can expect $N \gg Q$, however the magnitude of $Q$ will strongly depend on the number of species, $N_s$, involved in the chemical reactions and can even reach the order of thousands for more complex fuels \cite{lu2009toward}.

Combustion data sets can be obtained from numerical solutions of combustion models, or from experimental measurements of laboratory flames. When generating a numerical data set, we incorporate information about the chemistry of the process by selecting a chemical mechanism for combustion of a given fuel. The mechanism determines which chemical species are involved in the specified chemical reactions, and it can vary in complexity.

For the discussion of data sets, it is useful to introduce the distinction between premixed and non-premixed combustion. In the premixed case, fuel and oxidizer are first mixed before they are burned. In the non-premixed combustion, fuel and oxidizer originate from separate streams and have to first mix for the combustion to occur. In this latter case, the mixture fraction variable is an important scalar quantity that specifies the local fuel to oxidizer ratio and is defined as:
\begin{equation}\label{eq:mixture-fraction}
Z = \frac{\nu Y_{F} - Y_{O_2} + Y_{O_2,2}}{\nu Y_{F,1} + Y_{O_2,2}} ,
\end{equation}
for any point in space, where $Y_{F}$ is the local mass fraction of fuel, $Y_{O_2}$ is the local mass fraction of oxidizer in the mixture, and $\nu$ is their stoichiometric ratio \cite{Bilger1990}. $Y_{F,1}$ is the mass fraction of fuel in the fuel stream and $Y_{O_2,2}$ is the mass fraction of oxidizer in the oxidizer stream. The complete combustion of fuel happens at the stoichiometric mixture fraction $Z_{st}$. If $Z<Z_{st}$, the mixture is called lean (fuel deficient) and if $Z>Z_{st}$, the mixture is called fuel rich.

\begin{figure}
\includegraphics[width=\textwidth]{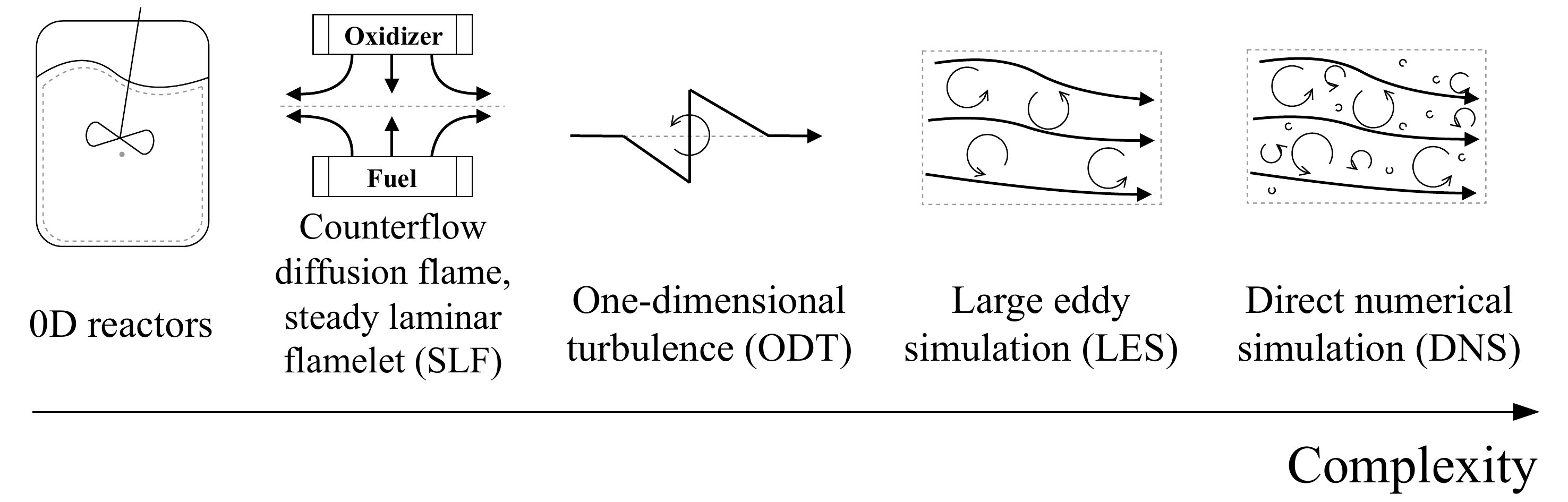}
\caption[Examples of common numerical combustion data sets schematically presented on the axis of increasing complexity.]
{Examples of common numerical combustion data sets schematically presented on the axis of increasing complexity.}
\label{fig:combustion-data-sets}
\end{figure}

Figure \ref{fig:combustion-data-sets} presents an overview of frequently encountered numerical data sets, ordered schematically by the amount of information about the combustion process they contain.
In particular:
\begin{itemize}
\item The zero-dimensional (0D) reactor model (e.g. perfectly stirred reactor (PSR)) assumes that combustion happens in a single point in space. This model can thus only include information about chemistry and thermodynamics of the combustion process evolving in time. It carries no information about spatial gradients of the thermo-chemical variables. The 0D reactor describes an ideal mixing process with $Da \ll 1$. When the data set is formed from the 0D reactor simulation, the $N$ rows of $\mathbf{X}$ are linked to points in time only.
\item The steady laminar flamelet (SLF) model \cite{peters1984laminar} assumes that fuel and oxidizer are two impinging streams, originating from the fuel and the oxidizer feed respectively. Combustion happens in infinitely thin sheets (flamelets) where fuel and oxidizer meet at varying stoichiometric proportions. The stoichiometry is specified by the mixture fraction variable, $Z$, and the thermo-chemical variables vary along the axis between pure oxidizer ($Z=0$) and pure fuel ($Z=1$). By varying the distance between the feeds (or the outlet velocities), a strain (specified by the strain rate $\chi$) is imposed on the flamelet, which can displace it from its equilibrium. The SLF model describes a non-premixed system with $Da \gg 1$. This is thus a two-parameter model with each row of $\mathbf{X}$ linked to one pair $(Z, \chi)$.
\item The counterflow diffusion flame (CFDF) model \cite{law2010combustion} is similar to the SLF model, as it assumes two separate streams of fuel and oxidizer diffusing into one another (non-premixed system). However, in contrast to SLF, the flow equations for CFDF are solved in the physical space. Flow is parameterized by the velocity gradient parameter $a$, which is an equivalent of the local strain rate. In the CFDF, each row of $\mathbf{X}$ is linked to a point in the physical space and time.
%\item The one-dimensional burner-stabilized premixed flame \cite{law2010combustion} is the premixed counterpart of the counterflow diffusion flame.
\item The one-dimensional turbulence (ODT) model \cite{kerstein1999one}, \cite{echekki2011one} can additionally incorporate the effects of turbulence by introducing eddy events on a one-dimensional domain. It allows for both spatial and temporal evolution of the flow. ODT can be used as a standalone model but can also serve as a subgrid model in large eddy simulations (LES).
\item The Reynolds-averaged Navier-Stokes (RANS) models \cite{ferziger2002computational} solve the time-averaged equations describing fluid dynamics and hence any sources of unsteadiness coming from turbulence are averaged.
\item The large eddy simulation (LES) \cite{ferziger2002computational} resolves large scales associated with the fluid dynamics processes but a subgrid model is required to account for processes occurring at the smallest scales. The choice for the subgrid model becomes particularly important in reactive flows, since combustion is inherently tied to the smallest scales.
\item The direct numerical simulation (DNS) \cite{ferziger2002computational} allows for the most accurate description of the coupled interaction between fluid dynamics and the thermo-chemistry since all subgrid processes are resolved directly. The scarcity of the DNS data sets is due to the large computational cost of performing DNS simulations, especially for more complex fuels.
\end{itemize}
More information on combustion theory and combustion models can be found in \cite{law2010combustion}, \cite{kee2005chemically}, or \cite{turns1996introduction}.

The selected data set is the starting point for applying a data science technique, and we often refer to those data sets as the \textit{training} data. From the description of various data sets presented above, it is visible that the choice of the data set will be pertinent to the type and quality of the analysis performed. For instance, we can expect that features identified will depend on the amount of information about the coupled phenomena of turbulence and chemistry that was initially captured in the data set.

\subsection{Data Preprocessing}

Combustion data sets typically contain variables of different numerical ranges (e.g. both temperature, which can range from hundreds up to thousands, and mass (or mole) fractions of chemical species, which take values in the range $Y_i \in \langle 0, 1 \rangle$). Scaling the data set is crucial in order to balance the importance of all state-space variables \cite{parente2013principal}.
The data set $\mathbf{X}$ can be centered and scaled as follows:
\begin{equation}\label{eq:pca-preprocessing}
\widetilde{\mathbf{X}} = (\mathbf{X} - \mathbf{\bar{X}}) \mathbf{D}^{-1} ,
\end{equation}
where $\mathbf{\bar{X}}$ contains the mean observations of each variable. $\mathbf{D}$ is the diagonal matrix of scales, where the $j$th element, $d_j$, from the diagonal is the scaling factor corresponding to the $j$th state-space variable, $\mathbf{X}_j$. A few of the common scaling criteria are collected in Table \ref{tab:scaling-criteria}.
The result is a centered and scaled data matrix, $\widetilde{\mathbf{X}}$. Other preprocessing means can include outlier detection or data sampling.
For the remainder of this chapter, we will assume that $\widetilde{\mathbf{X}}$ represents the matrix that has been adequately preprocessed.
\begin{table}
\caption[A few selected common data scaling criteria. $\sigma_j$ is the standard deviation of the $j$th state-space variable $\mathbf{X}_j$.]
{A few selected common data scaling criteria. $\sigma_j$ is the standard deviation, $max(\mathbf{X}_j)$ is the maximum value, $min(\mathbf{X}_j)$ is the minimum value, and $\bar{\mathbf{X}}_j$ is the mean observation of the $j$th state-space variable $\mathbf{X}_j$.}
\label{tab:scaling-criteria}
\begin{tabular}{@{}ll@{}}
\hline
Scaling technique & Scaling factor, $d_j$ \\
\hline
Auto  & $\sigma_j$ \\
Pareto  & $\sqrt{\sigma_j}$ \\
Range & $max(\mathbf{X}_j) - min(\mathbf{X}_j)$ \\
VAST & $\sigma_j^2 / \bar{\mathbf{X}}_j$ \\
\hline
\end{tabular}
\end{table}

\section{Feature Extraction Using Dimensionality Reduction Techniques}\label{section:feature}

Dimensionality reduction techniques offer a way to represent high-dimensional combustion data sets in a new, lower-dimensional basis. Such data representations are referred to as low-dimensional manifolds. Techniques such as PCA, NMF or autoencoders can extract those manifolds in an empirical way from the training data sets.
This approach belongs to the family of empirical low-dimensional manifolds (ELDMs) \cite{Pope_small_scales}, \cite{yang2013empirical}, and it is based on the idea that compositions occurring in combustion systems lie close to a low-dimensional manifold.
In addition, these techniques offer a way to extract meaningful features by exploiting the fact that the new basis can be better suited to represent certain physical phenomena underlying the original data. In this section, we review a few popular dimensionality reduction techniques, and we present their potential to detect features in combustion data sets.

\subsection{Description of the Analyzed Data Set}\label{subsection:DNS-data}

In this section, we apply various dimensionality reduction techniques on a numerical data set from a DNS simulation of a $CO$/$H_2$ turbulent jet \cite{sutherland2007quantitative} using the chemical mechanism consisting of 12 chemical species and 33 reactions \cite{yetter1991comprehensive}. Additional information regarding the numerical simulation can be found in \cite{sutherland2007quantitative}. The data set consists of a 2D slice extracted from the 3D domain arranged in a matrix $\mathbf{X}$. Each row of $\mathbf{X}$ corresponds to a point on the 2D grid, and therefore the matrix $\mathbf{X}$ can be thought of as a single time snapshot. Several such snapshots from different times in the simulation are available. The matrix $\mathbf{X}$ has 13 columns corresponding to temperature and 12 species' mass fractions\footnote{For the purpose of the feature extraction analysis presented here, all $N_s$ species were included in the data set.}.

\subsection{Extracting Features with Data Reduction Techniques}

\subsubsection{Principal Component Analysis} \label{subsubsec:PCA}

PCA \cite{jolliffe} projects the preprocessed data set $\widetilde{\mathbf{X}}$ onto a new basis represented by the orthonormal matrix of modes $\mathbf{A} \in \mathbb{R}^{Q \times Q}$. The matrix $\mathbf{A}$ can be obtained as the eigenvectors resulting from the eigendecomposition of the data covariance matrix $\mathbf{S} \in \mathbb{R}^{Q \times Q}$:
\begin{equation}\label{eq:pca-covariance}
\mathbf{S} = \frac{1}{N-1} \widetilde{\mathbf{X}}^\top \widetilde{\mathbf{X}} ,
\end{equation}
or from the singular value decomposition (SVD) of the data set:
\begin{equation}\label{eq:pca-svd}
\widetilde{\mathbf{X}} = \mathbf{U} \mathbf{\Sigma} \mathbf{A}^\top
\end{equation}
where $^\top$ denotes matrix transpose. This formulation makes PCA equivalent to proper orthogonal decomposition (POD) (see Chapter 6).
The preprocessed data matrix $\widetilde{\mathbf{X}}$ can then be projected onto $\mathbf{A}$ to obtain the principal components (PCs) matrix $\mathbf{Z} \in \mathbb{R}^{N \times Q}$:
\begin{equation}\label{eq:pca-data-transformation}
\mathbf{Z} = \widetilde{\mathbf{X}} \mathbf{A} 
\end{equation}
PCs represent the original observations represented in the new coordinate system defined by $\mathbf{A}$. It is worth noting that the columns of $\mathbf{Z}$ are linear combinations of the original state-space variables. The linear coefficients of that combination are specified in the entries of $\mathbf{A}$, and we will refer to them as \textit{weights}. Adopting the general notion of the data set as presented in eq.(\ref{eq:data-set}), the $j$th PC, $\mathbf{Z}_j$ (the $j$th column of $\mathbf{Z}$), is computed as:
\begin{equation}\label{eq:pca-first-pc}
\mathbf{Z}_j = \sum_{i = 1}^{Q} a_{ij} \widetilde{\mathbf{X}}_i = a_{1,j} \cdot \widetilde{T} + a_{2,j} \cdot \widetilde{p} + a_{3, j} \cdot \widetilde{Y}_1  + \dots + a_{Q,j} \cdot \widetilde{Y}_{N_s - 1} 
\end{equation}
where $a_{ij}$ are the elements (weights) from the $j$th column of $\mathbf{A}$ and tildes represent the preprocessing applied to each state-space variable. Since the basis matrix is orthonormal, $a_{ij} \in \langle -1, 1 \rangle$ and $\mathbf{A}^{-1} = \mathbf{A}^\top$.
By keeping a reduced number of the $q<Q$ first PCs, we obtain the closest\footnote{in terms of $L_2$, Frobenius, or trace norms, which follows from the Eckart-Young-Mirsky theorem \cite{eckart1936}.} rank-$q$ approximation of the original matrix:
\begin{equation}\label{eq:pca-approximation}
\mathbf{X} \approx \mathbf{X_q} = \mathbf{Z_q} \mathbf{A_q}^\top \mathbf{D} + \bar{\mathbf{X}} ,
\end{equation}
where the index $q$ denotes the truncation from $Q$ to $q$ components. Note that the reverse operation to the one defined in eq.(\ref{eq:pca-preprocessing}) has to be applied using matrices $\mathbf{D}$ and $\bar{\mathbf{X}}$.

We can assign physical meaning to the PCs by looking at the linear coefficients (weights) $a_{ij}$ from the basis matrix $\mathbf{A}$ \cite{parente2011investigation}, \cite{bellemans2018feature}. High absolute weight for a particular variable means that this variable is identified by PCA as important in the linear combination from the eq.(\ref{eq:pca-first-pc}). Moreover, since the vector $\mathbf{A}_j$ identifies the same span as $-\mathbf{A}_j$, only the relative sign of a particular weight with respect to the signs of other weights is important. In addition, PCs are ordered in a way that each PC captures more variance than the following one. Thus, we can expect the most important features identified by PCA to be visible in the first few PCs. This property of PCA can also guide the choice for the value $q$. Since PCs are decorrelated, increasing $q$ in the PCA reconstruction from eq.(\ref{eq:pca-approximation}) guarantees an improvement in the reconstruction errors.
\begin{figure}[h!]
\includegraphics[width=14pc]{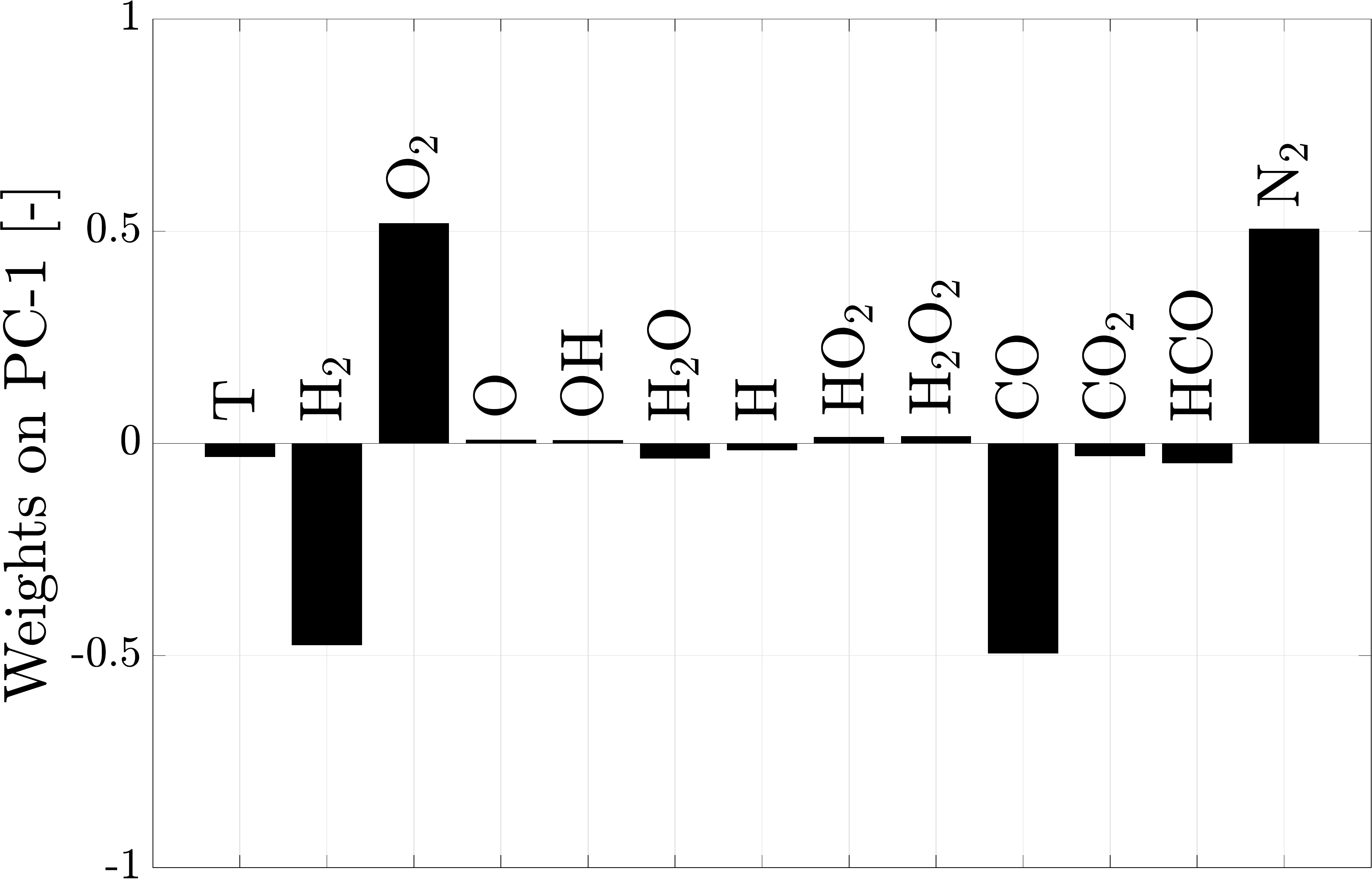}
\caption[The eigenvector weights $a_{ij}$ associated with the first PC found by PCA on a data set preprocessed with Range scaling.]
{The eigenvector weights $a_{ij}$ associated with the first PC found by PCA on a data set preprocessed with Range scaling.}
\label{fig:pc-1-range}
\end{figure}
\begin{figure}[h!]
\includegraphics[width=14pc]{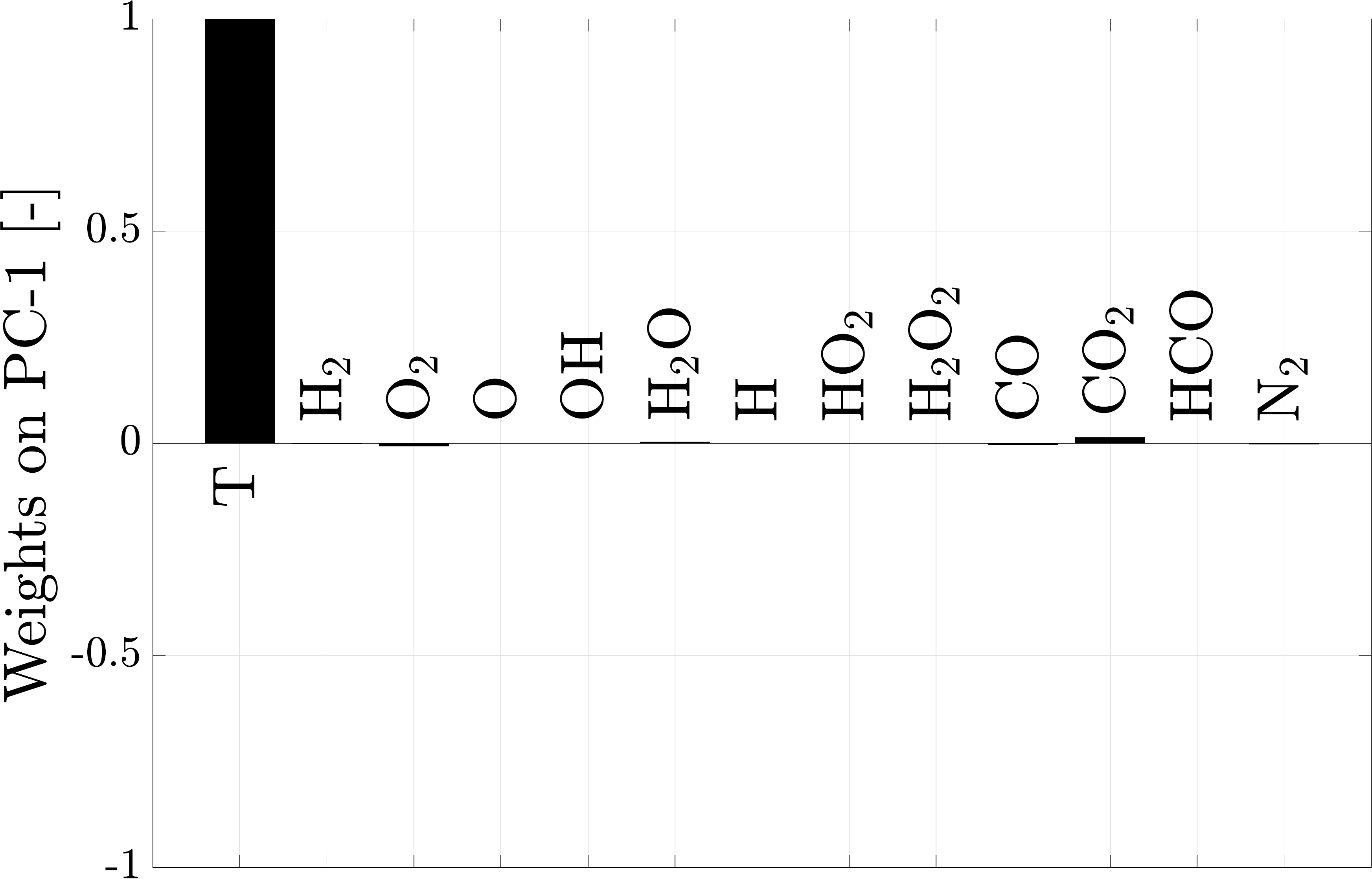}
\caption[The eigenvector weights $a_{ij}$ associated with the first PC found by PCA on a data set preprocessed with Pareto scaling.]
{The eigenvector weights $a_{ij}$ associated with the first PC found by PCA on a data set preprocessed with Pareto scaling.}
\label{fig:pc-1-pareto}
\end{figure}

Preprocessing the data set, prior to applying a dimensionality reduction technique, can have a significant impact on the shape of the low-dimensional manifold and on the types of features retrieved from the data set \cite{parente2013principal}, \cite{peerenboom2015dimension}. Figures \ref{fig:pc-1-range}-\ref{fig:pc-1-pareto} present the weights $a_{ij}$ associated with the first PC resulting from Range and Pareto scaling (see Table \ref{tab:scaling-criteria}) of the original data set.
Firstly, we observe that the structure of the PC can change significantly with the choice of the scaling technique.
If we further consider the mixture fraction variable as defined in eq.(\ref{eq:mixture-fraction}) as a linear combination of fuel $Y_{F}$ and oxidizer $Y_{O_2}$ mass fractions, then it can be observed that the coefficients in front of $Y_{F}$ and $Y_{O_2}$ are of opposite signs in that definition.
Using Range scaling (Figure \ref{fig:pc-1-range}), the first PC can be attributed to the mixture fraction variable, where the only high weights are for the fuel ($H_2$ and $CO$) and oxidizer components ($O_2$ and $N_2$), and the two have opposite signs. The correlation between the mixture fraction variable and the first PC is $99.96\%$.
It is worth noting here, that the mixture fraction was not among the variables in the original data set $\mathbf{X}$ and PCA identifies it in an automatic way.
With Pareto scaling (Figure \ref{fig:pc-1-pareto}), the first PC is almost entirely aligned with the temperature variable and carries almost no information about the mass fractions of the chemical species.

\begin{figure}[h!]
\includegraphics[width=14pc]{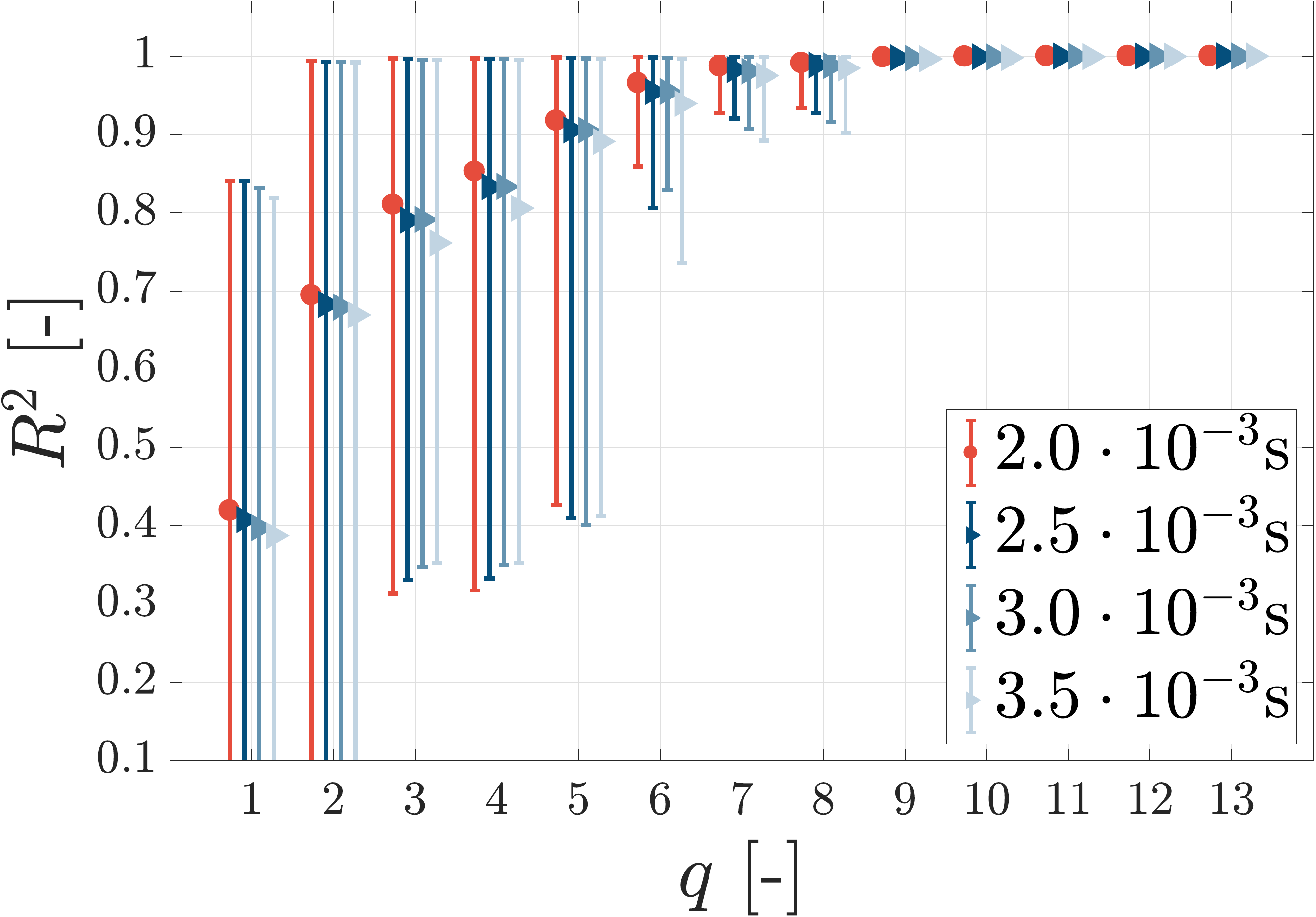}
\caption[Mean, minimum (worst reconstruction), and maximum (best reconstruction) $R^2$ values for reconstructing three snapshots separated by $\Delta t = 5$ms (blue triangles) using modes from the snapshot at time $t = 2.0 \cdot 10^{-3}$s (red circles). Shown for increasing $q$ in the PCA reconstruction.]
{Mean, minimum (worst reconstruction), and maximum (best reconstruction) $R^2$ values for reconstructing three snapshots separated by $\Delta t = 5$ms (blue triangles) using modes from the snapshot at time $t = 2.0 \cdot 10^{-3}$s (red circles). Shown for increasing $q$ in the PCA reconstruction.}
\label{fig:DNS2-r2}
\end{figure}

\begin{figure}[h!]
\includegraphics[width=14pc]{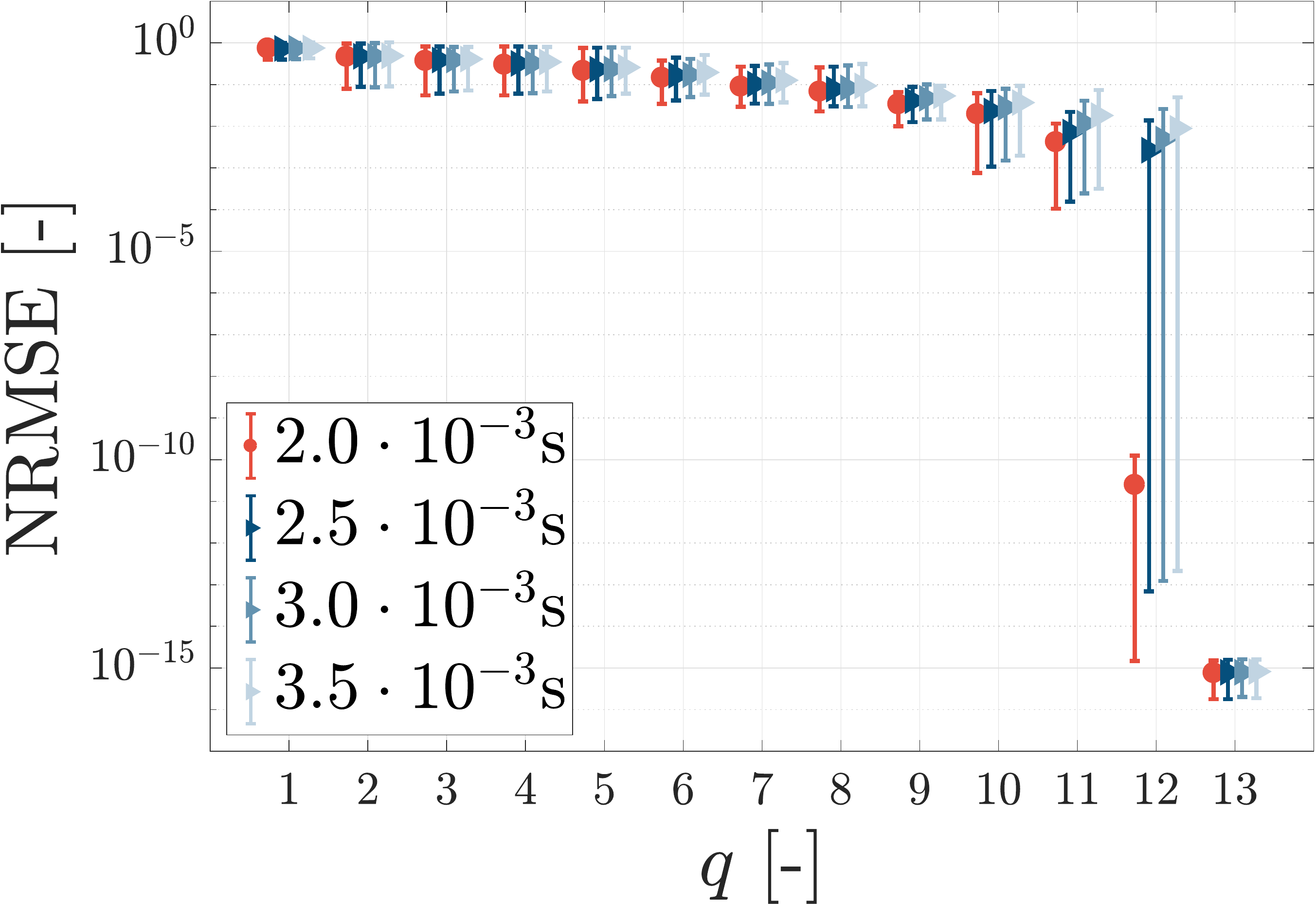}
\caption[Mean, minimum (best reconstruction), and maximum (worst reconstruction) NRMSE values for reconstructing three snapshots separated by $\Delta t = 5$ms (blue triangles) using modes from the snapshot at time $t = 2.0 \cdot 10^{-3}$s (red circles). Shown for increasing $q$ in the PCA reconstruction.]
{Mean, minimum (best reconstruction), and maximum (worst reconstruction) NRMSE values for reconstructing three snapshots separated by $\Delta t = 5$ms (blue triangles) using modes from the snapshot at time $t = 2.0 \cdot 10^{-3}$s (red circles). Shown for increasing $q$ in the PCA reconstruction.}
\label{fig:DNS2-nrmse}
\end{figure}

In a previous study \cite{Biglari_Suther_CF_2012}, the authors have demonstrated that PCA can identify PCs that are independent of the filter width on a fully resolved jet flame. PCA was performed on the state-space variables filtered using a top-hat filter of varying widths. To test the capability of PCA to extract time-invariant features of the data set, we can also use a fixed set of modes to reconstruct new, unseen data, such as data from future time steps of the same temporally evolving system. If the reconstruction process leads to errors that are comparable with the ones obtained for the training data, we can expect that PCA captures the \textit{essence} of the physical processes underlying the system. We demonstrate this using 2D slices from the DNS data set from four time intervals separated by $\Delta t = 5$ms. In Figures \ref{fig:DNS2-r2}-\ref{fig:DNS2-nrmse}, three future data snapshots (blue triangles) are reconstructed using $q$ first eigenvectors found on the initial snapshot (red circles). Figure \ref{fig:DNS2-r2} shows the coefficients of determination, which can be computed for the $j$th variable $\mathbf{X}_j$ in the data set as:
\begin{equation}\label{eq:r2}
R_j^2 = 1 - \frac{\sum_{i=1}^N (x_{ij} - \hat{x}_{ij})^2}{\sum_{i=1}^N (x_{ij} -  \bar{x}_j)^2}
\end{equation}
where $x_{ij}$ is the $i$th observation of the $j$th variable, $\hat{x}_{ij}$ is the PCA reconstruction of that observation, and $\bar{x}_j$ is the average observation of $\mathbf{X}_j$. The coefficient of determination $R^2$ measures the \textit{goodness of the model fit} with respect to fitting the data with the mean value $\bar{x}_j$. Values $R^2 \in ( - \infty, 1 \rangle$, where $R^2=1$ means a perfect fit. The smaller the $R^2$ value gets, the worse the model fit.
Figure \ref{fig:DNS2-nrmse} shows the normalized root-mean-squared errors (NRMSE) on a logarithmic scale. NRMSE can be computed for the $j$th variable $\mathbf{X}_j$ in the data set as:
\begin{equation}\label{eq:nrmse}
\text{NRMSE}_j = \frac{1}{\bar{x}_j} \cdot \sqrt{\frac{\sum_{i=1}^{N} (x_{ij} - \hat{x}_{i,j})^{2}}{N}}
\end{equation}
In Figures \ref{fig:DNS2-r2}-\ref{fig:DNS2-nrmse}, the markers represent the mean values of $R^2$ or NRMSE averaged over all variables in the data set. The bars range from the minimum and the maximum value achieved for any variable in the data set.
It can be observed that for all the reconstructed snapshots, the error metrics show comparable values. This indicates the capability of PCA to capture generalized, time-invariant features of temporally evolving systems. It can also be observed that the mean errors grow for future snapshots, which indicates that there might be a limit on the time separation $\Delta t$ for which we can extend the applicability of the features found.

\subsubsection{Local Principal Component Analysis}
\label{section:local_pca}

\begin{figure}
\includegraphics[width=\textwidth]{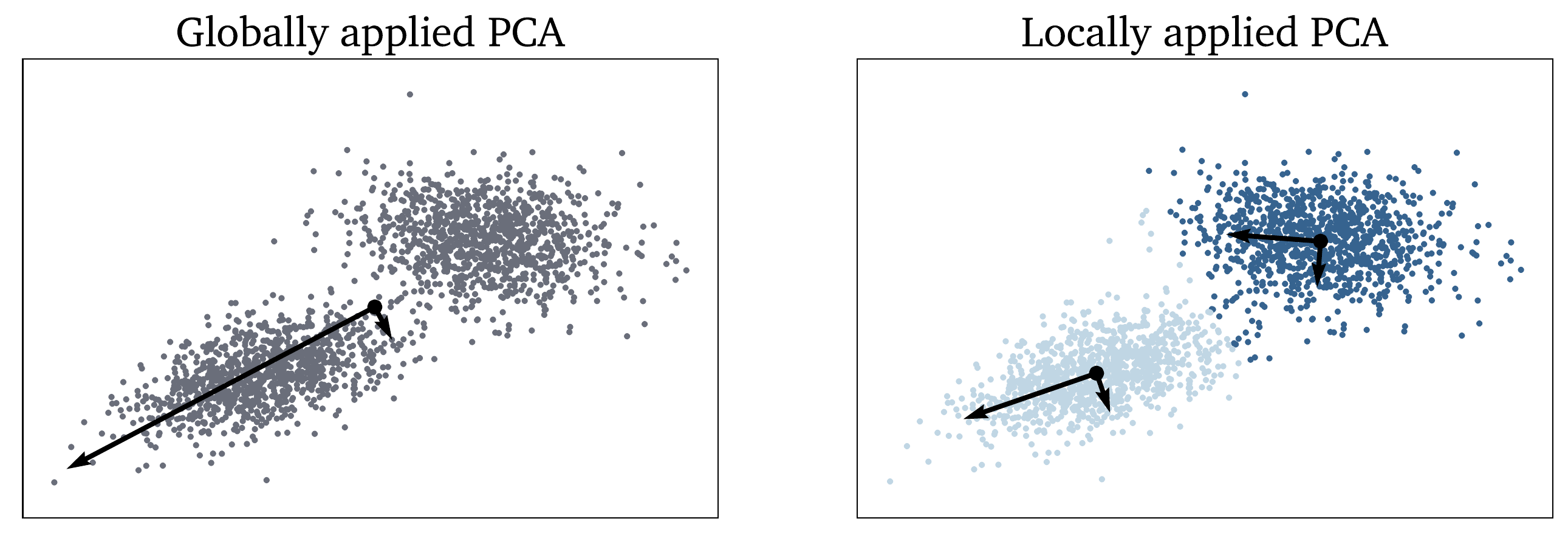}
\caption[Schematic distinction between global and local PCA on a synthetic 2D data set. The arrows represent the two eigenvectors from the matrix $\mathbf{A}$.]
{Schematic distinction between global and local PCA on a synthetic 2D data set. The arrows represent the two global/local modes from the matrix $\mathbf{A}$, defining the directions of the largest variance in the global/local data.}
\label{fig:global-local-pca}
\end{figure}

PCA can also be applied locally (LPCA), on portions of the original data set. In LPCA, the clustering algorithm is first applied to the data set to partition observations into local clusters. Next, PCA is performed separately in each cluster. LPCA can not only allow for further reduction in dimensionality by adjusting to the potential nonlinearities of the data set, but it can also aid in detecting local features of the data. Local PCA can also guide the decision on how to partition the data into clusters, since it grants control of the data reconstruction errors. Figure \ref{fig:global-local-pca} shows the difference between the global and the local approach on a synthetic data set that is visibly composed of two distinct clusters. In the global case, PCA is performed on the entire data set, ignoring the apparent clusters. When the data set is partitioned and PCA is performed locally in each cluster, a different set of eigenvectors and PCs is tied to each identified cluster of data. Those local PCs can better represent the physics or the underlying features of a particular cluster by capturing the local variance instead of the global one.
\begin{figure}
\includegraphics[width=\textwidth]{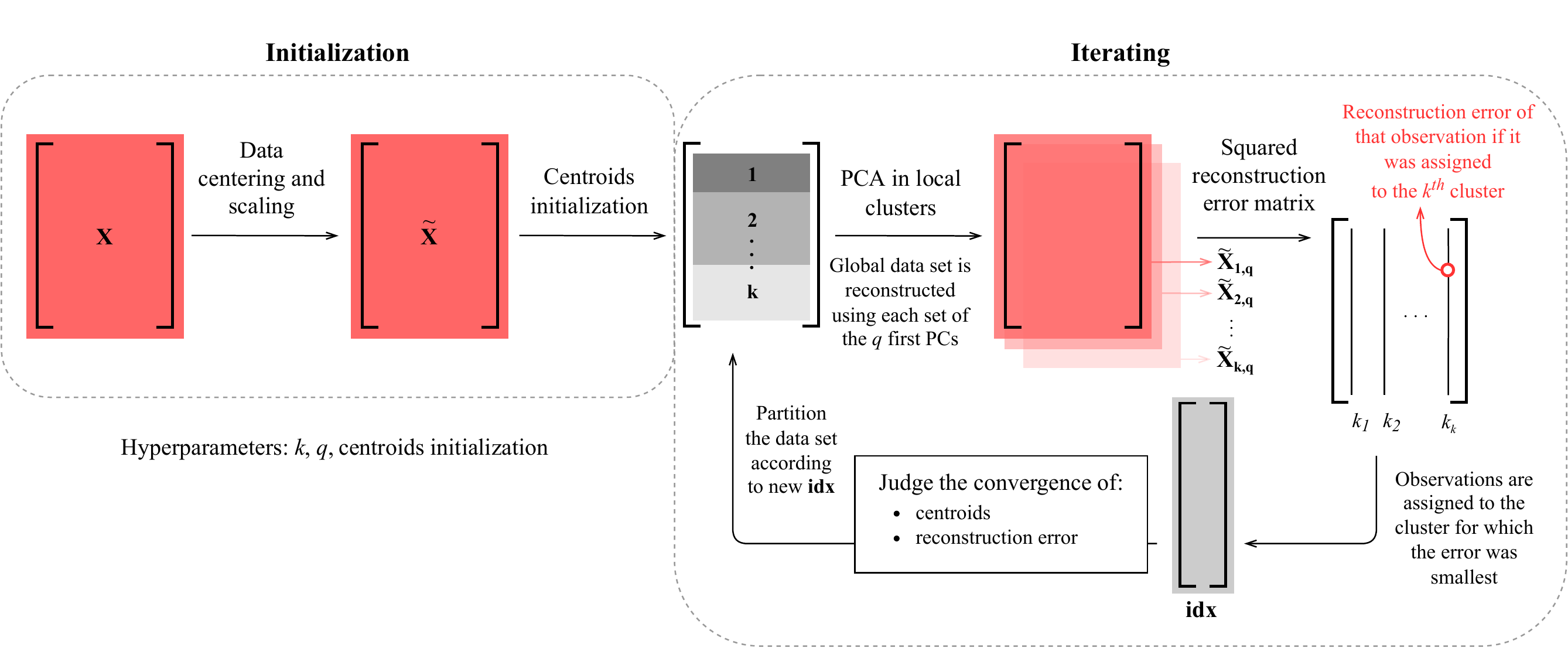}
\caption[Diagram presenting schematically the VQPCA algorithm.]
{Diagram presenting schematically the VQPCA algorithm.}
\label{fig:vqpca}
\end{figure}
In LPCA, the $i$th observation $\mathbf{x}_i$ is reconstructed from the local, $q$-dimensional manifold in an analogous way to eq.(\ref{eq:pca-approximation}):
\begin{equation}\label{eq:rec_VQPCA}
\mathbf{x}_i \approx \mathbf{x}_{\mathbf{q}, i} = \mathbf{z}_{\mathbf{q}, i} ( \mathbf{A}_{\mathbf{q}}^{(k)} )^\top \mathbf{D} + \mathbf{c}^{(k)}
\end{equation}
where $k$ is the index of the cluster to which the $i$th observation belongs, $\mathbf{z}_{\mathbf{q}, i}$ is the $i$th observation represented in the local, truncated basis $\mathbf{A}_{\mathbf{q}}^{(k)}$ identified on the $k$th cluster. Each cluster is centered separately using the centroid $\mathbf{c}^{(k)}$ of the $k$th cluster and typically the global diagonal matrix of scales $\mathbf{D}$ is applied in each cluster.

Data clustering, prior to applying local PCA, can be performed with any algorithm of choice. One of the techniques discussed in this chapter is the vector quantization PCA (VQPCA) algorithm \cite{kambhatla1997dimension}, \cite{parente2009identification}, presented schematically in Figure \ref{fig:vqpca}. This is an iterative algorithm in which the observations are assigned to the cluster for which the local PCA reconstruction error of that observation is the smallest. The hyperparameters of the algorithm include: the number of clusters $k$ to partition the data set, the number of PCs $q$ used to approximate the data set at each iteration, and the initial cluster partitioning. The latter is pre-defined by setting the initial cluster centroids. The most straightforward way is to initialize centroids randomly, but another viable option is to use partitioning resulting from a different clustering technique such as K-Means \cite{macqueen1967some}. The algorithm iterates until convergence of the centroids and of the reconstruction error is reached. The reconstruction error is measured between the centered and scaled data set $\widetilde{\mathbf{X}}$ and the approximation $\mathbf{\widetilde{X}}_{\mathbf{i,q}}$ using the $q$ first PCs computed from the eigenvectors found on the $i$th cluster.
More details on the VQPCA algorithm can be found in \cite{parente2009identification} and \cite{parente2011investigation}.

Another possible way of clustering the combustion data sets is to use a conditioning variable, such as mixture fraction, and partition the observations based on bins of that variable. If the mixture fraction is used, observations are first split into fuel-lean and fuel-rich parts at the stoichiometric mixture fraction $Z_{st}$. If more than two clusters are requested, lean and rich sides can then be further divided. The approach of performing local PCA on clusters identified through binning the mixture fraction vector is referred to as FPCA.

\begin{figure}
\includegraphics[width=14pc]{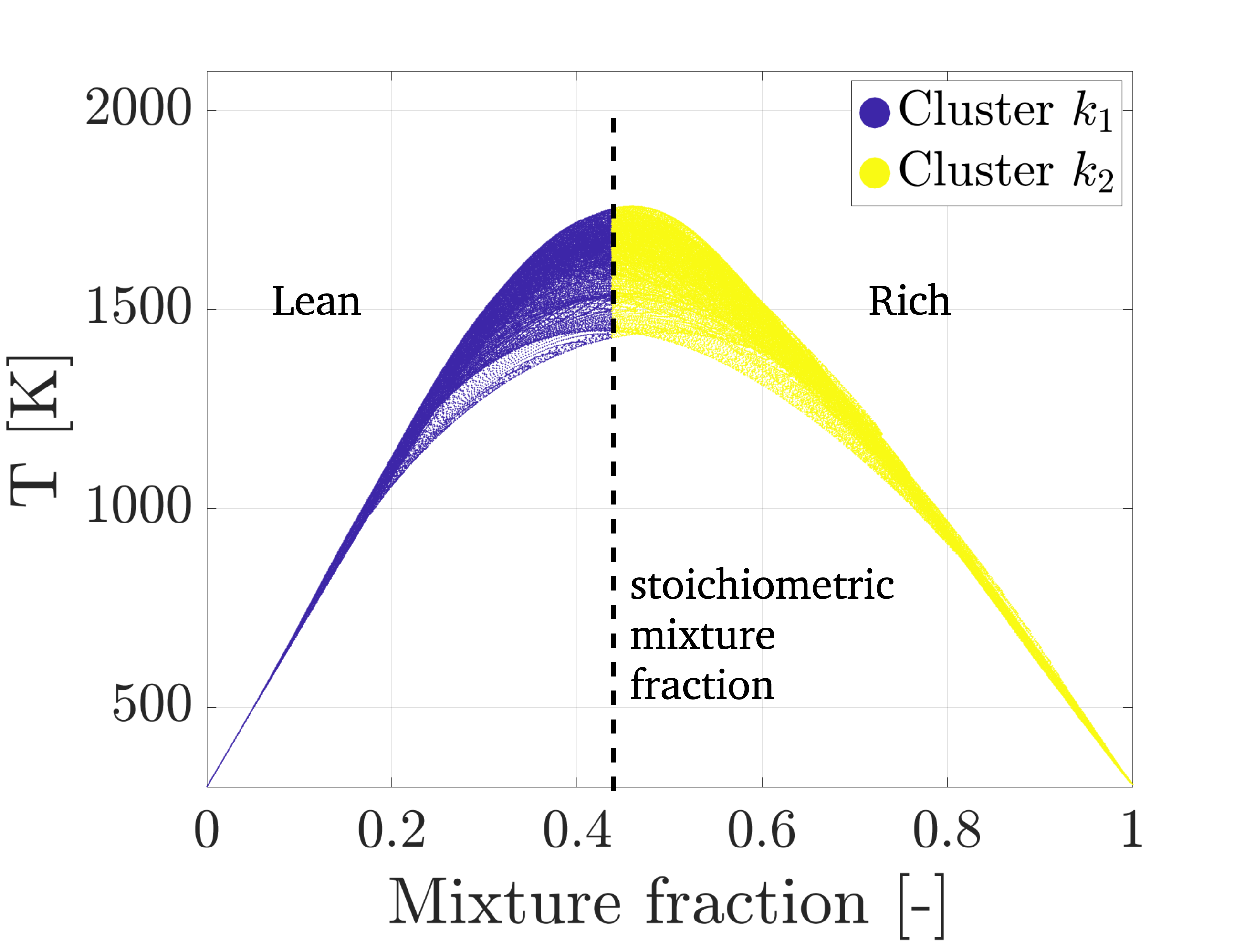}
\caption[Conditioning variable partitioning.]
{Conditioning variable partitioning into $k=2$ clusters in the mixture fraction-temperature space.}
\label{fig:lpca-mf-bins}
\end{figure}

\begin{figure}
\includegraphics[width=14pc]{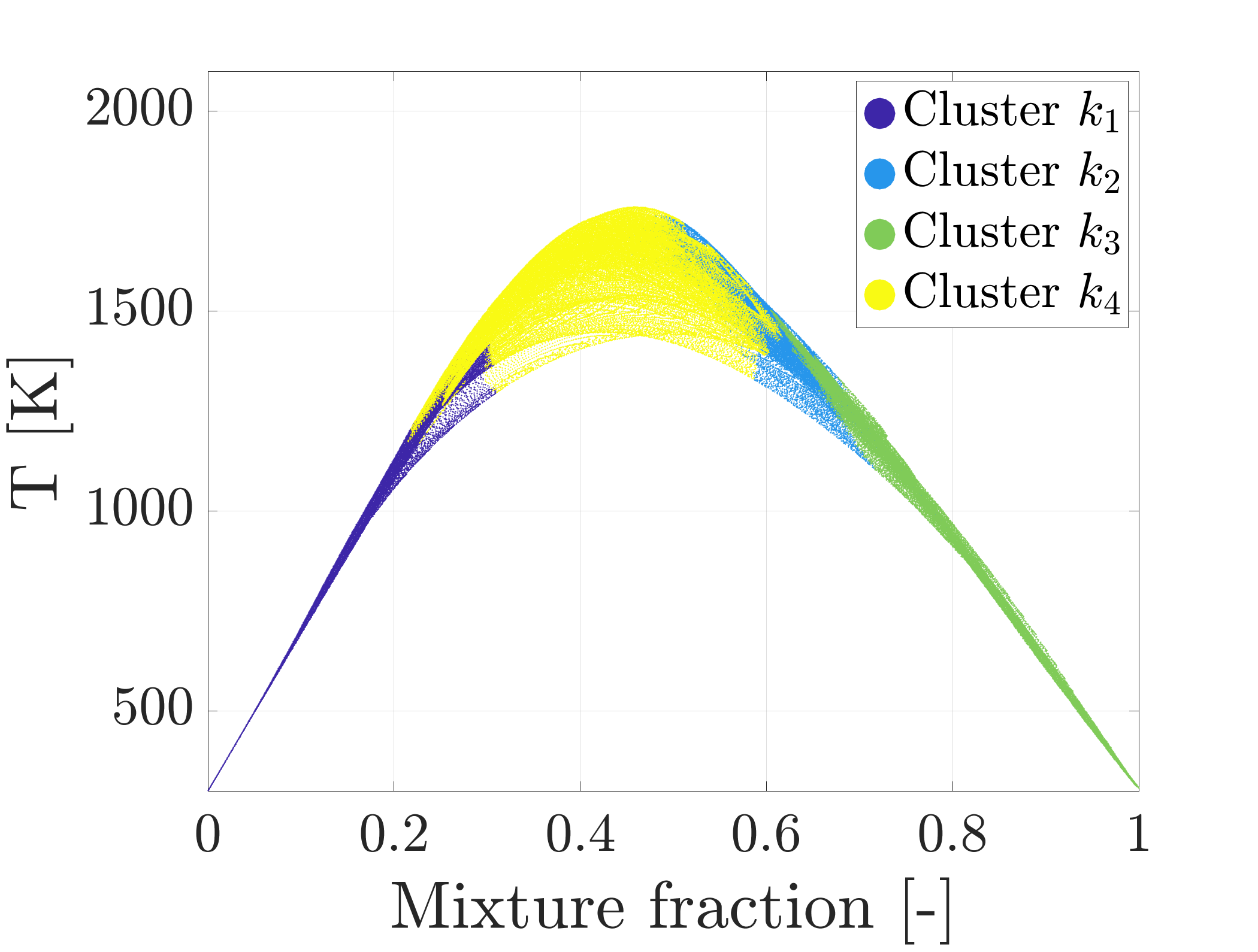}
\caption[VQPCA partitioning.]
{VQPCA partitioning into $k=4$ clusters in the mixture fraction-temperature space.}
\label{fig:lpca-vqpca}
\end{figure}

\begin{figure}
\includegraphics[width=\textwidth]{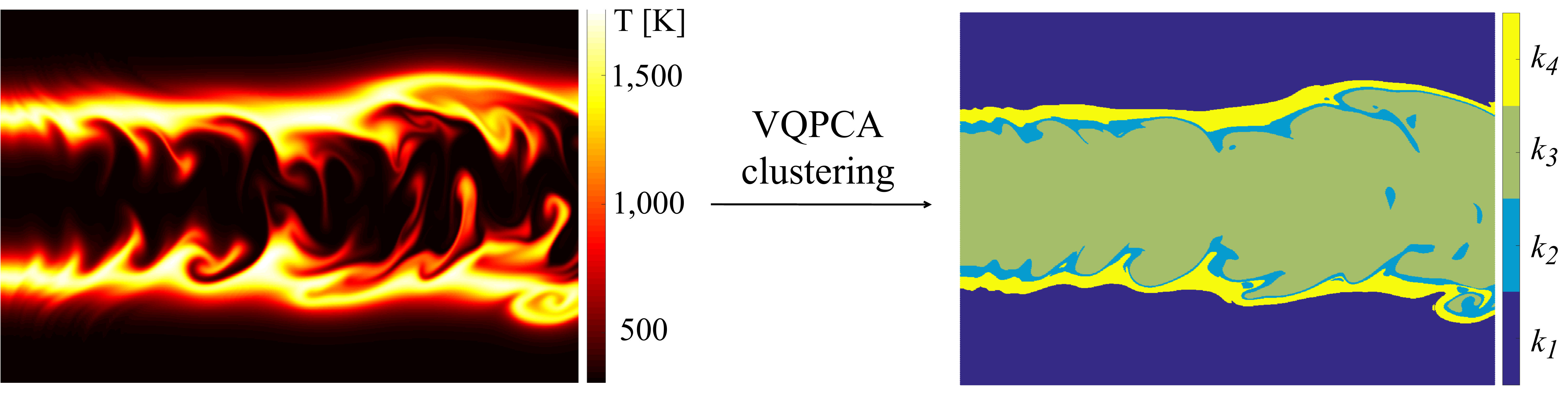}
\caption[Temperature profile of the DNS data set (left) and the result of partitioning the data set into $k=4$ clusters using the VQPCA algorithm (right).]
{Temperature profile of the DNS data set (left) and the result of partitioning the data set into $k=4$ clusters using the VQPCA algorithm (right).}
\label{fig:clustering-dns1}
\end{figure}

Local PCA was investigated on the benchmark DNS data set (\S\ref{subsection:DNS-data}) using two clustering algorithms. Figures \ref{fig:lpca-mf-bins}-\ref{fig:lpca-vqpca} show a comparison between two clustering results in the space of temperature and the mixture fraction variable. Figure \ref{fig:lpca-mf-bins} presents clustering into $k=2$ clusters using bins of mixture fraction as the conditioning variable. This partitioning can be thought of as \textit{hardcoded} in a sense that the split into two clusters will always be performed at $Z_{st}$. The features retrieved on local portions of data can thus only be attributed to the lean and rich zones. In contrast, Figure \ref{fig:lpca-vqpca} presents clustering into $k=4$ clusters using VQPCA algorithm. The VQPCA method could distinguish between the oxidizer (cluster $k_1$), the fuel (cluster $k_3$), and the region where the two meet close to stoichiometric conditions (clusters $k_2$ and $k_4$). This is even more apparent if we plot the result of the VQPCA clustering on a spatial grid in Figure \ref{fig:clustering-dns1}. The space is clearly divided into the inner fuel jet ($k_3$), the outer oxidizer layer ($k_1$), and the two thin reactive layers ($k_2$, $k_4$) for which the temperature is the highest.

The success of local PCA in extracting features depends on the clustering technique used. In a previous study \cite{d2020impact}, VQPCA has been compared to other clustering algorithms, and better results in terms of clustering quality and algorithm speed have been obtained. An unsupervised clustering algorithm based on the VQPCA partitioning has recently been proposed \cite{d2020unsupervised} to perform data mining on a high-dimensional DNS data set. If an algorithm such as VQPCA is used, the types of features found depend on data preprocessing and can additionally depend on the hyperparameters. By changing the number of clusters or the number of PCs in the approximation, the user can potentially retrieve different features. This has been also investigated in a previous study \cite{d2020analysis} for a more complex, high-dimensional DNS data set. This is in contrast with techniques such as binning on the conditioning variable, for which we might have an \textit{a priori} clarity for what the features will represent.

\subsubsection{Non-negative Matrix Factorization}

NMF \cite{paatero1994positive} is an algorithm for factorizing data matrices whose elements are non-negative. This technique can thus be applied to the data set as defined in eq.(\ref{eq:data-set}) since the thermo-chemical state-space variables are non-negative, provided that at the preprocessing step the data set is scaled but not centered ($\widetilde{\mathbf{X}} = \mathbf{X} \mathbf{D}^{-1}$). That is to ensure that we are not introducing negative elements in the matrix $\widetilde{\mathbf{X}}$ through centering by mean values\footnote{Alternatively, one can also subtract minimum values from each variable, thus making the range of each state-space variable start at 0.} as is done in eq.(\ref{eq:pca-preprocessing}). Given the non-negative data set $\mathbf{X} \in \mathbb{R}^{N \times Q}$, the scaling matrix $\mathbf{D}$ and the value for $q$, NMF aims to find two non-negative matrices $\mathbf{W} \in \mathbb{R}^{N \times q}$ and $\mathbf{F} \in \mathbb{R}^{q \times Q}$ such that:
\begin{align}
\mathbf{X} \approx \mathbf{X_q} = \mathbf{W} \mathbf{F} \mathbf{D}
\end{align}
The matrix of non-negative factors $\mathbf{F}$ can be regarded as the one containing a basis (analogous to the matrix $\mathbf{A}$ found by PCA). The matrix $\mathbf{W}$ represents the compressed data, namely the NMF scores and is thus analogous to the PCs matrix $\mathbf{Z}$.
The factorization to $\mathbf{W}$ and $\mathbf{F}$ is not unique, and various optimization algorithms exist \cite{lin2007projected}, \cite{berry2007algorithms}. In this chapter, we use the MATLAB\textsuperscript{\tiny\textregistered} routine \texttt{nnmf}, which minimizes the root-mean-squared residual \cite{berry2007algorithms}:
\begin{align}
D = \frac{1}{\sqrt{N \cdot Q}} ||\mathbf{X} - \mathbf{W} \mathbf{F}||_F ,
\end{align}
starting with random initial values for $\mathbf{W}$ and $\mathbf{F}$, where the subscript $F$ denotes the Frobenius norm. This optimization might reach local minimum and thus repeating the algorithm can yield different factorizations.

\begin{figure}
\includegraphics[width=14pc]{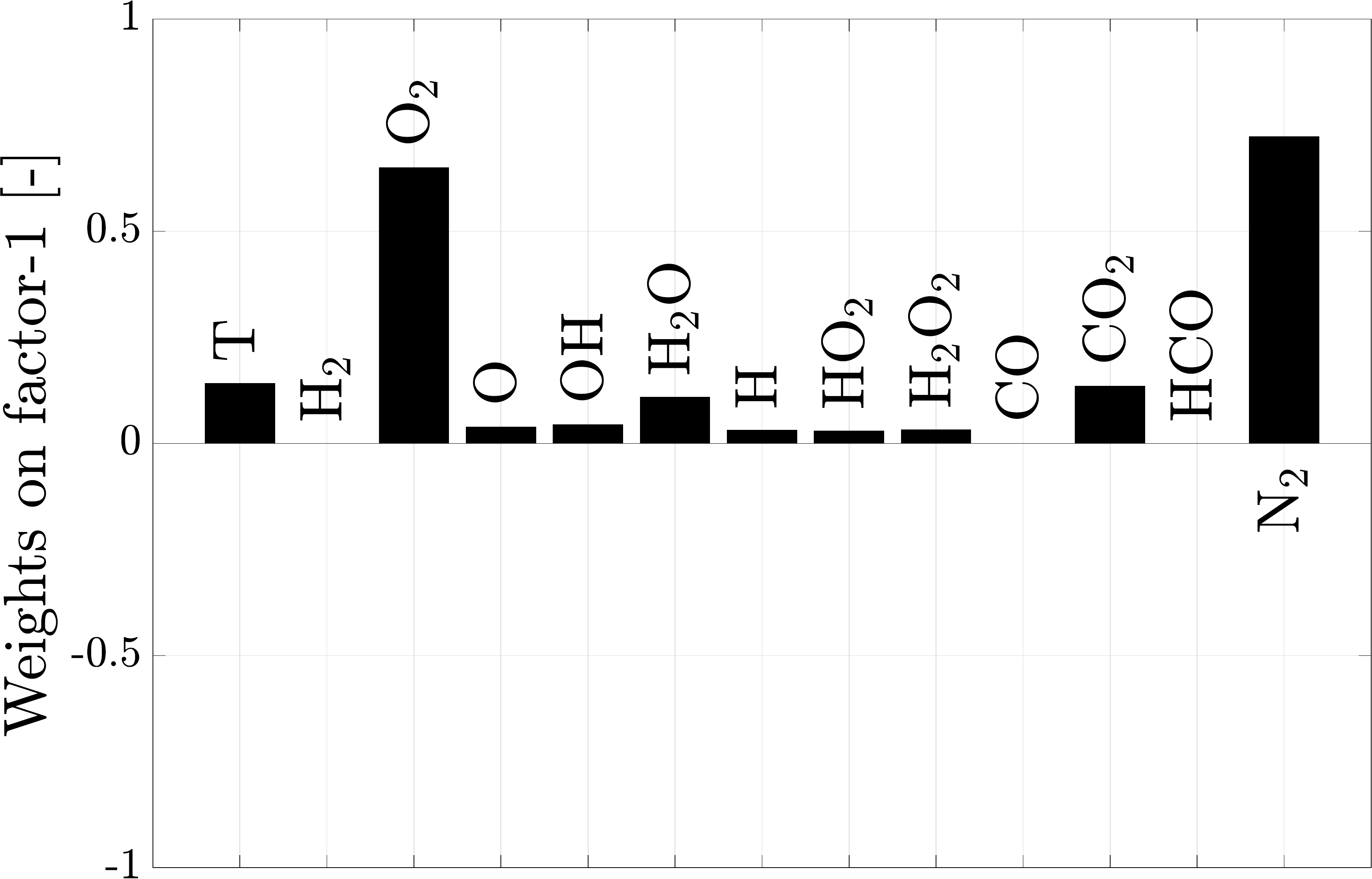}
\caption[First non-negative factor $\mathbf{f}_1$. NMF performed on Range-scaled data with $q=2$.]
{First non-negative factor $\mathbf{f}_1$. NMF performed on Range-scaled data with $q=2$.}
\label{fig:nmf-factor-1}
\end{figure}

\begin{figure}
\includegraphics[width=14pc]{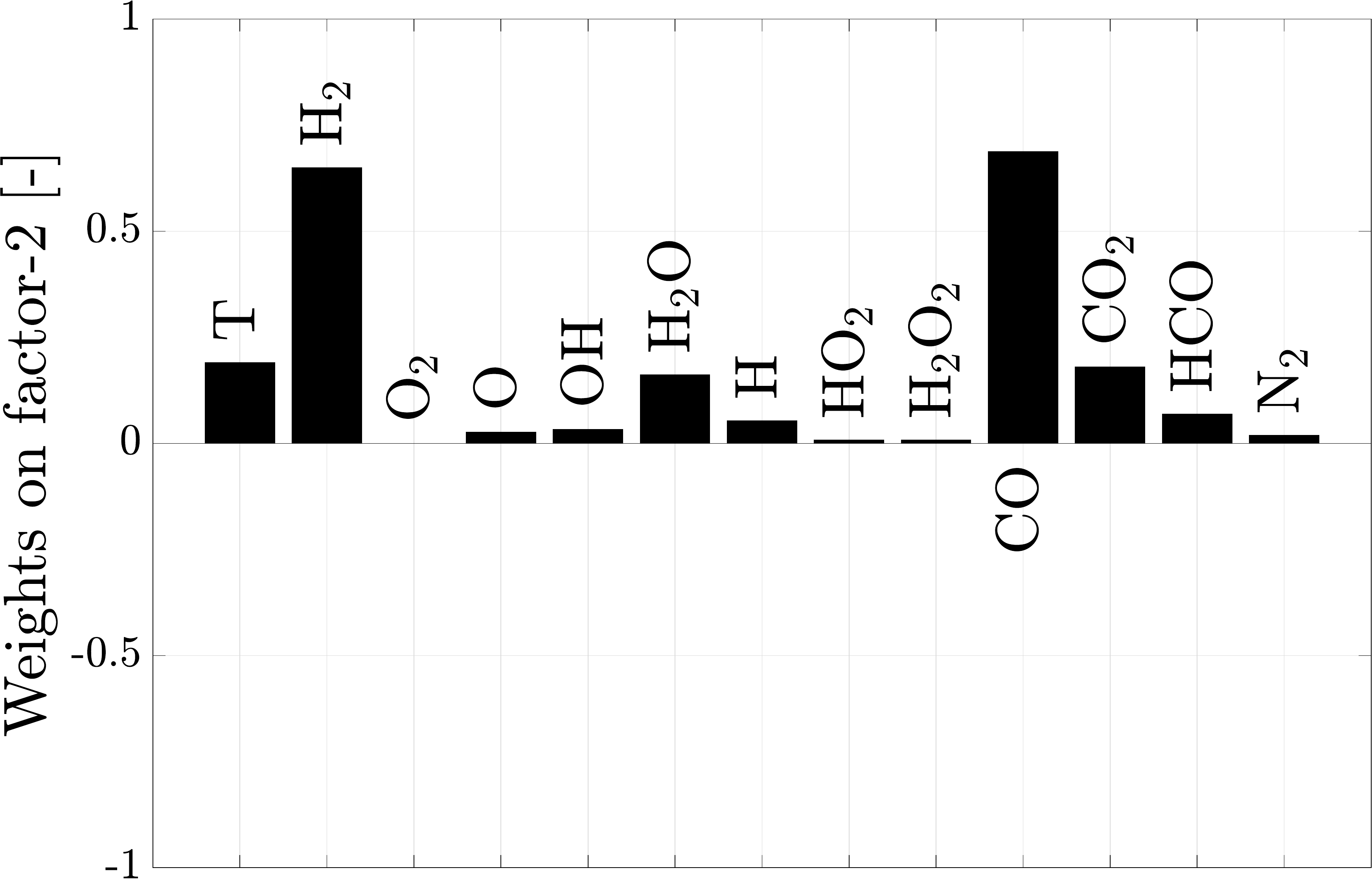}
\caption[Second non-negative factor $\mathbf{f}_2$. NMF performed on Range-scaled data with $q=2$.]
{Second non-negative factor $\mathbf{f}_2$. NMF performed on Range-scaled data with $q=2$.}
\label{fig:nmf-factor-2}
\end{figure}

Similarly as was done in PCA, we can look at the non-negative factor weights (the elements of $\mathbf{F}$) to assign physical meaning to the factors. With the non-negative constraint on $\mathbf{F}$, only non-negative weights can be found. Figures \ref{fig:nmf-factor-1}-\ref{fig:nmf-factor-2} show the first two non-negative factors that together represent the mixture fraction variable (compare with Figure \ref{fig:pc-1-range}). The oxidizer components are included in the first factor and the fuel components in the second. If NMF is compared to PCA, the latter can be thought of as more robust since NMF required two modes to capture the same information (the mixture fraction variable) as was included in a single PCA mode.

\subsubsection{Autoencoders}

The autoencoder \cite{goodfellow2016deep}, \cite{wang2016auto} is a type of an unsupervised artificial neural network (ANN) whose aim is to learn the $q$-dimensional representation (\textit{embedding}) of the $Q$-dimensional data set such that the reconstruction error between the input and the output layer is minimized. The standard form of an autoencoder is the feedforward neural network having an input layer and an output layer with the same number of neurons, and one or more hidden layers. Given one hidden layer, the \textit{encoding} process takes as an input the preprocessed data matrix $\widetilde{\mathbf{X}} \in \mathbb{R}^{N \times Q}$ and maps it to $\mathbf{H} \in \mathbb{R}^{N \times q}$, with $q < Q$:
\begin{align}
\mathbf{H} = f(\widetilde{\mathbf{X}} \mathbf{G} + \mathbf{B})
\end{align}
where the columns of $\mathbf{H}$ are referred to as the \textit{codes} and $f$ is the activation function such as sigmoid, rectified linear unit (ReLU), or squared exponential linear unit (SELU). $\mathbf{G} \in \mathbb{R}^{Q \times q}$ is the matrix of weights and $\mathbf{B} \in \mathbb{R}^{N \times q}$ is the matrix of biases.
At the \textit{decoding} stage, $\mathbf{H}$ is mapped to the reconstruction $\widetilde{\mathbf{X}}_{\mathbf{q}}$:
\begin{align}
\widetilde{\mathbf{X}} \approx \widetilde{\mathbf{X}}_{\mathbf{q}} = f'(\mathbf{H} \mathbf{G'} + \mathbf{B'})
\end{align}
where $f'$, $\mathbf{G'}$, and $\mathbf{B'}$ may be unrelated to $f$, $\mathbf{G}$, and $\mathbf{B}$. The encoding/decoding process can thus be summarized as:
\begin{align}
\widetilde{\mathbf{X}} \in \mathbb{R}^{N \times Q} \xrightarrow{\text{encoding}} \mathbf{H} \in \mathbb{R}^{N \times q} \xrightarrow{\text{decoding}} \widetilde{\mathbf{X}}_{\mathbf{q}} \in \mathbb{R}^{N \times Q}
\end{align}

\begin{figure}
\includegraphics[width=14pc]{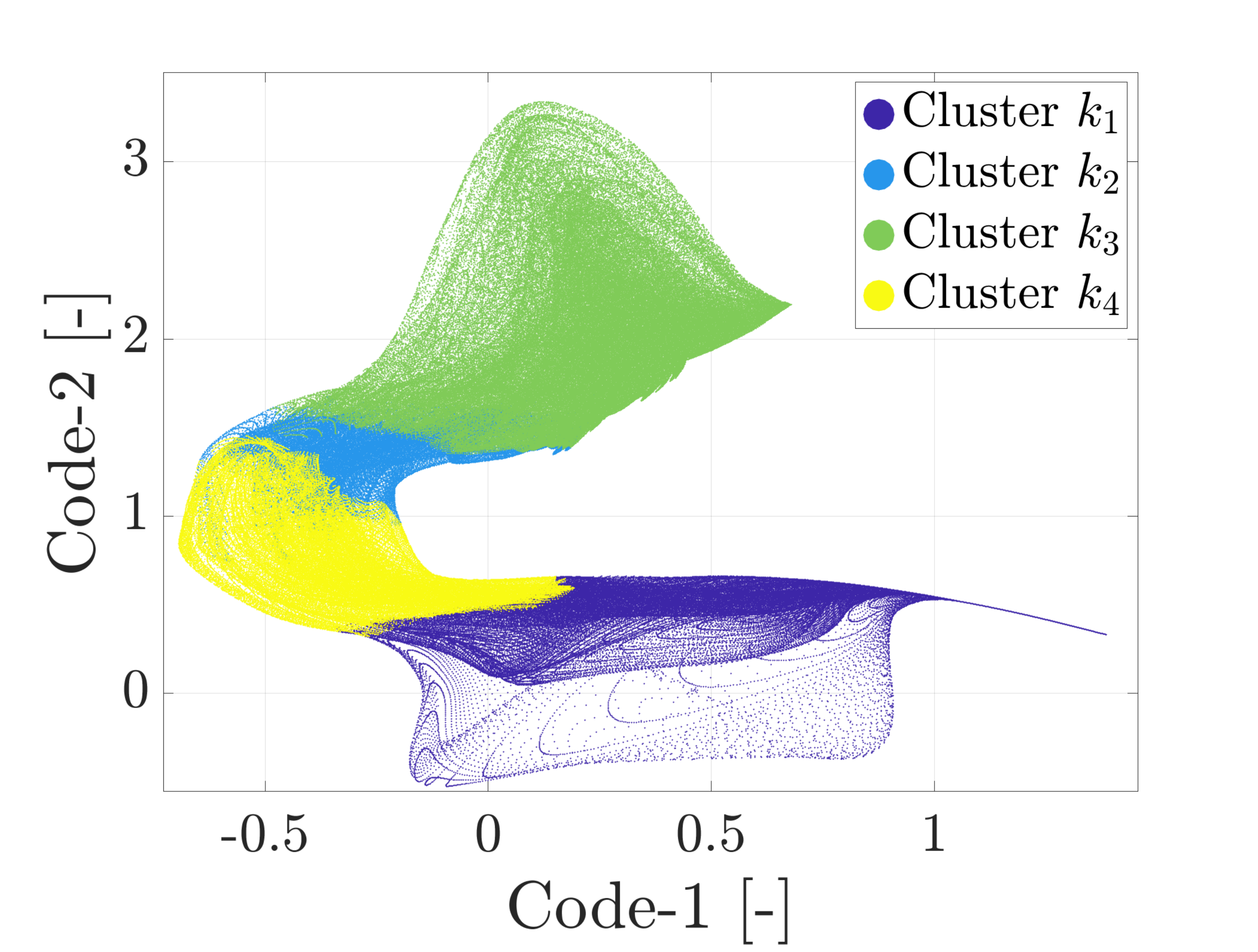}
\caption[2D manifold obtained using an Autoencoder with five hidden layers and SELU activation function. Colored by the result of VQPCA partitioning into $k=4$ clusters.]
{2D manifold obtained using an Autoencoder with five hidden layers and SELU activation function. Colored by the result of VQPCA partitioning into $k=4$ clusters.}
\label{fig:autoencoder-VQPCA}
\end{figure}

In this section, we use an autoencoder with five hidden layers and SELU activation function and generate a 2D embedding ($q=2$) of the original data set. Figure \ref{fig:autoencoder-VQPCA} shows the 2D manifold obtained after the autoencoder compression (represented by the matrix $\mathbf{H}$). The manifold is colored by the previously obtained result of partitioning via the VQPCA algorithm. From Figure \ref{fig:autoencoder-VQPCA}, it can be observed that the result of VQPCA partitioning still uniformly divides the autoencoder manifold. Clusters $k_1$ and $k_3$, representing the oxidizer and fuel respectively, are located at the opposing ends of the manifold. Thus, it is possible to think of that manifold as describing the progress of the combustion process, with the fuel and oxidizer meeting in the center ($k_2$ and $k_3$) where they finally react.

\subsubsection{Linear Operations That Aid in Feature Interpretation} \label{sec:linear-operations}

Several linear operations can aid in the interpretation of PCs or low-dimensional manifolds. One of such techniques is the rotation of modes/factors \cite{abdi2003factor}, \cite{bellemans2018feature} with the varimax orthogonal rotation \cite{kaiser1958varimax} used most commonly. In the varimax rotation, we first select which $n$ factors should be rotated together. It then maximizes the sum of variances of the squared weights within each of the $n$ rotated factors:
\begin{align}
\mathcal{V} = \sum (w_{ij}^2 - \bar{w}_{ij}^2)^2
\end{align}
where $w_{ij}$ is the $i$th weight on the $j$th factor and $\bar{w}_{ij}^2$ is the mean of the squared weights \cite{abdi2003factor}.
After rotation, those $n$ factors explain the same total amount of variance as they did together before the rotation, but variance is now redistributed differently among the selected factors. Varimax-rotated factors typically have high weights on fewer variables than the original factors, and that can aid in their physical interpretation. The rotated factors can then be used as the new basis to represent the original data set. Varimax and several other rotation methods are available within the MATLAB\textsuperscript{\tiny\textregistered} routine \texttt{rotatefactors}.

\begin{figure}[h!]
\includegraphics[width=14pc]{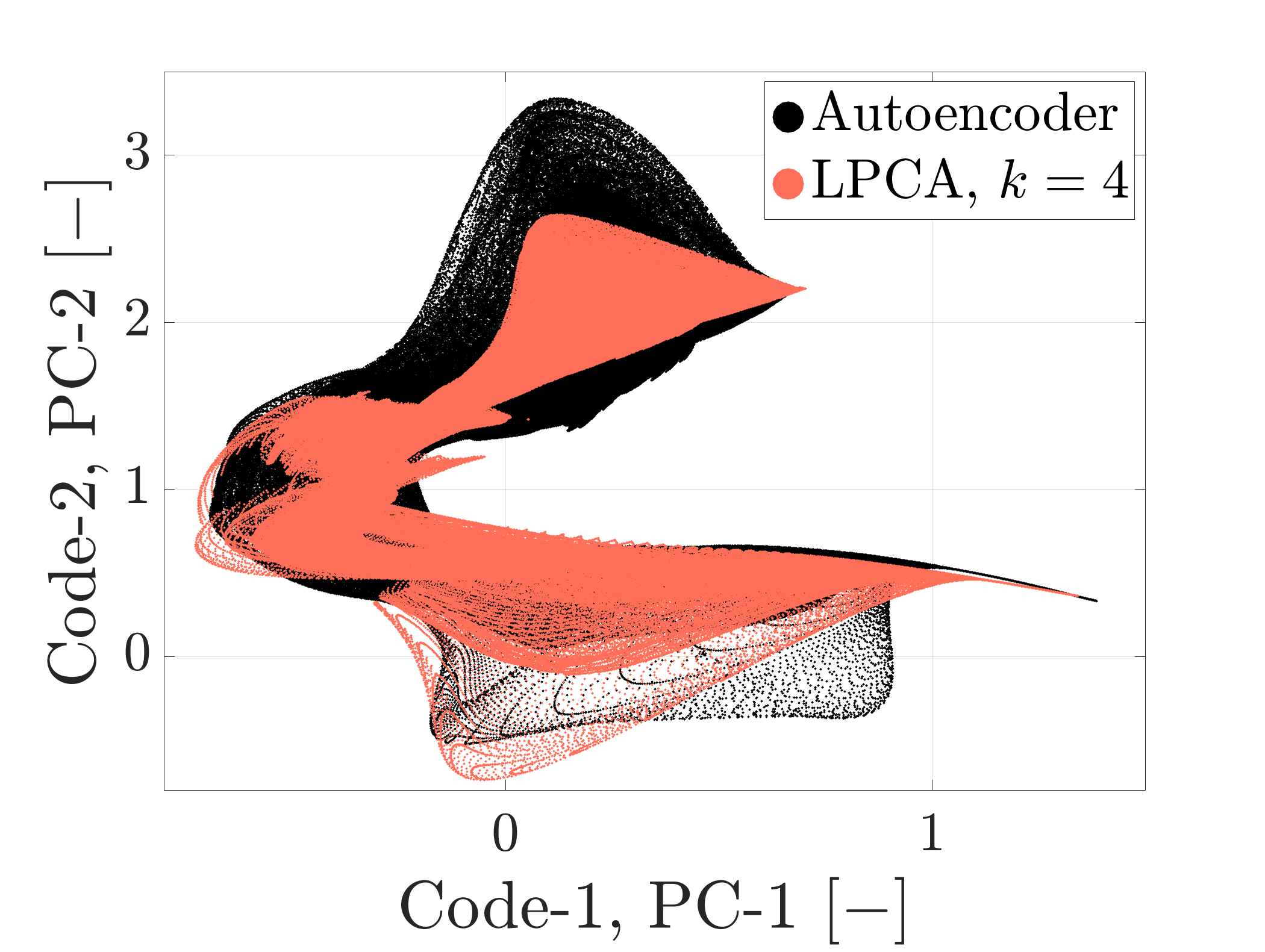}
\caption[The 2D manifold obtained from local PCA using VQPCA clustering algorithm and increasing the number of clusters, overlayed with the autoencoder manifold using the Procrustes analysis. VQPCA performed with $k=4$ clusters.]
{The 2D manifold obtained from local PCA using VQPCA clustering algorithm and increasing the number of clusters, overlayed with the autoencoder manifold using the Procrustes analysis. VQPCA performed with $k=4$ clusters.}
\label{fig:procrustes-4}
\end{figure}

\begin{figure}[h!]
\includegraphics[width=14pc]{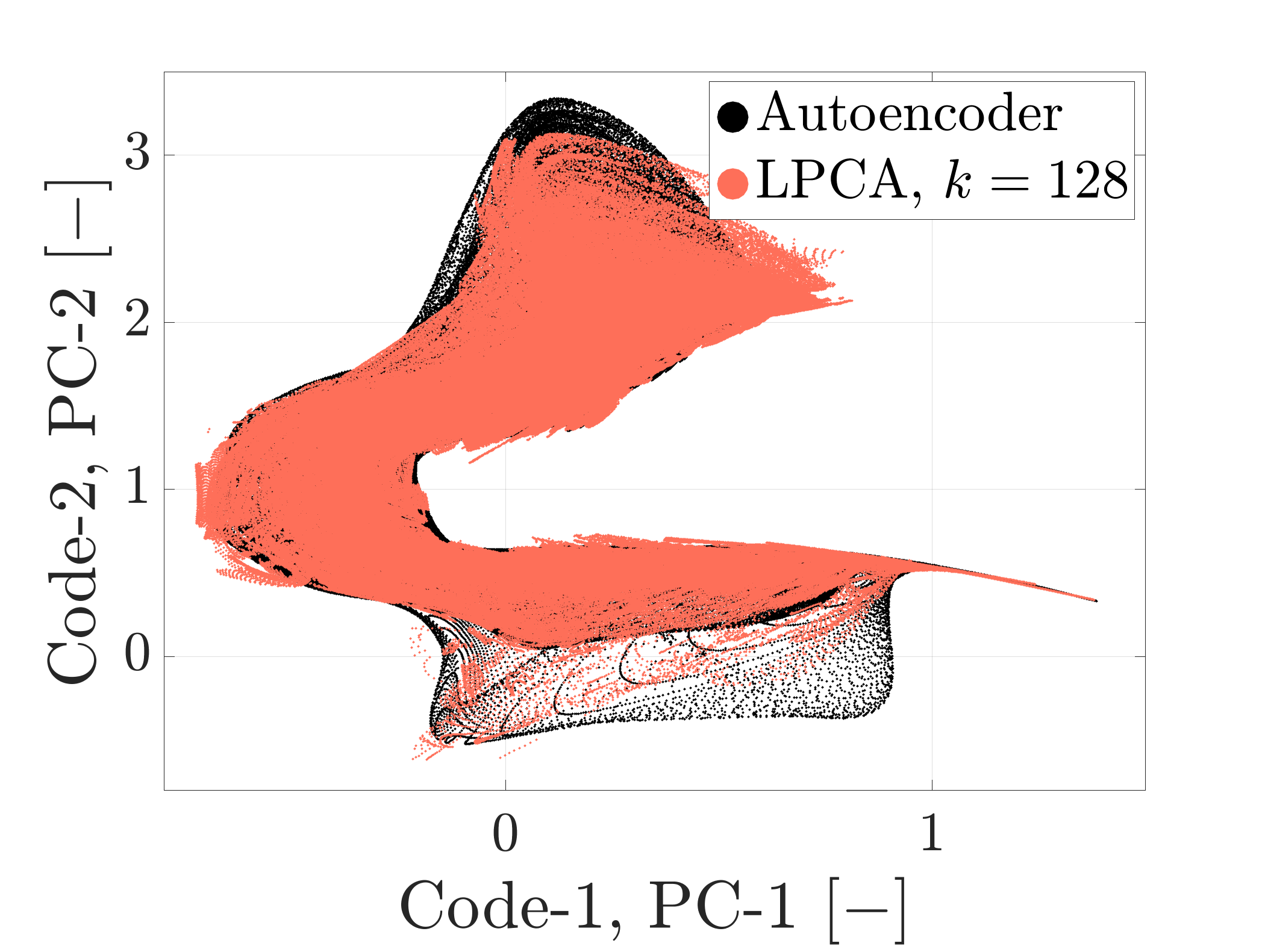}
\caption[The 2D manifold obtained from local PCA using VQPCA clustering algorithm and increasing the number of clusters, overlayed with the autoencoder manifold using the Procrustes analysis. VQPCA performed with $k=128$ clusters.]
{The 2D manifold obtained from local PCA using VQPCA clustering algorithm and increasing the number of clusters, overlayed with the autoencoder manifold using the Procrustes analysis. VQPCA performed with $k=128$ clusters.}
\label{fig:procrustes-128}
\end{figure}

Another interesting technique is the Procrustes analysis \cite{seber2009multivariate}, which is a series of linear operations that allow translation, rotation, or scaling of the low-dimensional manifold. This can be particularly useful when manifolds obtained from two dimensionality reduction techniques should be compared. Figures \ref{fig:procrustes-4}-\ref{fig:procrustes-128} present the Procrustes transformation (using the MATLAB\textsuperscript{\tiny\textregistered} routine \texttt{procrustes}) of the LPCA manifold onto an autoencoder manifold. With the series of linear transformations the manifolds overlay each other, and the match between them becomes closer to exact as the number of clusters with which VQPCA was performed is increased. This suggests that the data reconstruction error for the 2D approximation from LPCA will approach the one from autoencoder decoding when the number of clusters is high enough.


\section{Transport of Principal Components}\label{section:pc_transport}

In \S\ref{section:data-sets}, we have seen that the number of thermo-chemical state-space variables $Q$ determines the original dimensionality of the data set. This number also reflects how many transport equations for the state-space variables should be solved in a numerical simulation. The general transport equation for the set $\mathbf{\Phi} = \mathbf{X}^\top = [T, p, Y_1, Y_2, \dots, Y_{N_s-1}]^\top$ of state-space variables is:
\begin{equation}\label{eq:species_transport}
\rho \frac{D \mathbf{\Phi}}{D t} = - \nabla \cdot ( \mathbf{j_{\Phi}} ) + \mathbf{S_{\Phi}}
\end{equation}
where $\mathbf{j_{\Phi}}$ is the mass-diffusive flux of $\mathbf{\Phi}$ relative to the mass-averaged velocity, and $\mathbf{S_{\Phi}}$ is the volumetric rate of production of $\mathbf{\Phi}$ (also referred to as the source of $\mathbf{\Phi}$). Performing detailed simulations with large chemical mechanisms with significant number of chemical species is still computationally prohibitive. In \cite{sutherland2009combustion}, the authors proposed to use PCA to reduce the number of transport equations that solve a combustion process. Instead of solving the original set of $Q$ transport equations, the original variables are first transformed to the new basis identified by PCA on the training data set $\mathbf{X}$. Next, the truncation from $Q$ to $q$ first PCs is performed. Transport equations for the $q$ first PCs can be formulated from
eq.(\ref{eq:species_transport}) using the truncated basis matrix $\mathbf{A_q}$:
\begin{equation}\label{eq:PC_transport}
\rho\frac{D \mathbf{z}}{D t} = - \nabla \cdot ( \mathbf{j_{z}} ) + \mathbf{S_{z}}
\end{equation}
where $\mathbf{z} = \mathbf{Z_q}^\top$ (with $\mathbf{Z_q} = (\mathbf{X} - \mathbf{\bar{X}}) \mathbf{D}^{-1} \mathbf{A_q}$), $\mathbf{j_{z}} = \mathbf{A_q} \mathbf{j_{\Phi}}$, and $\mathbf{S_{z}} = \mathbf{A_q} \mathbf{S_{\Phi}} \mathbf{D}^{-1}$ (also referred to as the PC-sources). This is the discrete analog of the Galerkin projection methods described in Chapters 6 and 14. Note that the source terms of the PCs, $\mathbf{S_{z}}$, are scaled (but not centered) using the same scaling matrix $\mathbf{D}$ as applied on the data set $\mathbf{X}$.

The challenge associated with the resolution of the PC-transport equation is related to the PC-source terms. The latter are highly nonlinear functions (based on Arrhenius expressions) of the state space variables. The nonlinearity of the chemical source terms strongly impacts the degree of reduction attainable using the projection of the species transport equations onto the PCA basis \cite{Isaac_2014_CF}, \cite{Isaac2015_CF}. A solution to this problem is to use PCA to identify the most appropriate basis to parameterize the ELDM and then, both the thermo-chemical state-space variables and the PC-source terms can be nonlinearly regressed onto the new basis. This allows to overcome the shortcomings associated with the multi-linear nature of PCA and to reduce the number of components required for an accurate description of the state-space.

The construction of an appropriate low-dimensional manifold requires training data. This might be seen as a limitation of the approach since all system states are required before applying model reduction. Although initial studies on PCA models involved DNS data of turbulent combustion \cite{sutherland2009combustion}, \cite{Pope_small_scales}, recent studies have demonstrated \cite{Biglari_Suther_CF_2015}, \cite{Echekki_CF_2015}, \cite{Coussement_CF_2016}, \cite{malik2018principal}, \cite{dalakoti2020priori} that PCA-based models can be trained on simple and inexpensive systems, such as zero-dimensional reactors and one-dimensional flames (see \S\ref{section:data-sets}), and then applied to model complex systems, such as the flame-vortex interaction \cite{Coussement_MG-local-PCA_2013}, the flame-turbulence interaction \cite{Isaac2015_CF} as well as the turbulent premixed \cite{Coussement_CF_2016} and non-premixed flames \cite{malik2020combustion}. Up-to-date, the PC-transport approach has been demonstrated for a wide range of problems involving ideal reactors, including batch and PSR reactors.

\subsection{Non-linear Regression Models}

The thermo-chemical state-space variables $\mathbf{\Phi}$ and the PC-source terms $\mathbf{S_{Z_q}}$ can be mapped to the $q$-dimensional PC space using a nonlinear regression function $\mathscr{F}$, such that:
\begin{equation*}
\phi \approx \mathscr{F}( \mathbf{Z_q} )
\end{equation*}
where $\phi$ represents any of the dependent variables ($T, p, Y_1, Y_2, \dots, Y_{N_s-1}$, or $\mathbf{S_{Z_q}}$).
The models used in the literature to generate the function $\mathscr{F}$ include, but are not limited to:
\begin{itemize}
\item multivariate adaptive regression splines (MARS) \cite{MARS},
\item artificial neural networks (ANN) \cite{ANN},
\item Gaussian process regression (GPR) \cite{GPR_Rasmussen}.
\end{itemize}

In a previous study \cite{Isaac2015_CF}, the authors compared different regression models in their ability to accurately map highly nonlinear functions (such as the chemical source terms) to the PCA manifold.
In this section, we will focus on the use of GPR for state-space and source term parameterization. The choice of GPR is due to its semi-parametric nature that increases the generality of the approach. GPR employs Gaussian mixtures to capture information about the relation between data and input parameters, making predictions of non-observed system states more reliable than in fully parametric approaches.

\subsection{Validation of the PC-Transport Approach in LES Simulations}

To demonstrate the application of the PC-transport approach, we show the results of the LES of a multi-dimensional flame \cite{malik2020combustion}. A piloted methane-air diffusion flame (referred to as Flame D) \cite{Flame_D} for which high-fidelity experimental data is available is selected. Flame D is fueled by a mixture of $CH_{4}$ and air ($25\%\ /\ 75\%$ by volume), at $Re=22,400$ and 294K, respectively. The fuel jet is surrounded by pilot flames that stabilize the flame. Finally, an air co-flow surrounds the flame.

The development of a reduced-order model for subsequent application in LES requires the availability of training data. The most critical aspect when generating a training data set is to make sure that the generated state-space includes all the possible states accessed during the actual simulation. At the same time, that data cannot come from  expensive simulations. Parametric calculations on inexpensive canonical configurations can provide an effective solution in this context, although the nature of the flame archetype chosen is an open subject of discussion. Considering the non-premixed nature of Flame D, it was decided to rely on unsteady one-dimensional laminar counter diffusion flames. The use of unsteady inlet velocity profiles allows to explore transient flame behavior, including extension and re-ignition phenomena. The calculations were performed using the OpenSMOKE++ suite developed in Politecnico di Milano \cite{OpenSmoke}. The GRI 3.0 \cite{GRI30} mechanism, involving 35 species and 253 reactions (excluding $NO_x$), was used. The inlet conditions were set equal to the experimental ones \cite{Flame_D}. Multiple simulations were carried out by varying the strain rate, from equilibrium to complete extinction. The unsteady solutions were saved on a uniform grid of 400 points over a $0.15$ m domain. All of the unsteady data from the various simulations was used collectively for the PCA analysis. The final data set consisted of $\sim80,000$ observations for each of the state-space variables.

\begin{table}
\caption[The elements $a_{ij}$ of the basis matrix $\mathbf{A}$ identified by PCA when the data set was composed of five major chemical species only.]
{The elements $a_{ij}$ of the basis matrix $\mathbf{A}$ identified by PCA when the data set was composed of five major chemical species only.}
\label{Table_A_5}
\begin{tabular}{@{}cccccc@{}}
\hline
Species & $\mathbf{A}_1$ & $\mathbf{A}_2$ & $\mathbf{A}_3$ & $\mathbf{A}_4$ & $\mathbf{A}_5$\\
\hline
$H_{2}O$ & -0.02 & 0.51 & 0.45 & -0.73 & 0.02\\
$O_{2}$ & -0.18 & -0.67 & -0.22 & -0.60 & 0.30\\
$N_{2}$ & -0.64 & -0.01 & -0.15 & -0.09 & -0.74\\
$CH_{4}$ & 0.73 & -0.14 & -0.20 & -0.27 & -0.56\\
$CO_{2}$ & -0.02 & 0.50 & -0.82 & -0.14 & 0.21\\
\hline
\end{tabular}
\end{table}
\begin{figure}[h!]
\includegraphics[width=14pc]{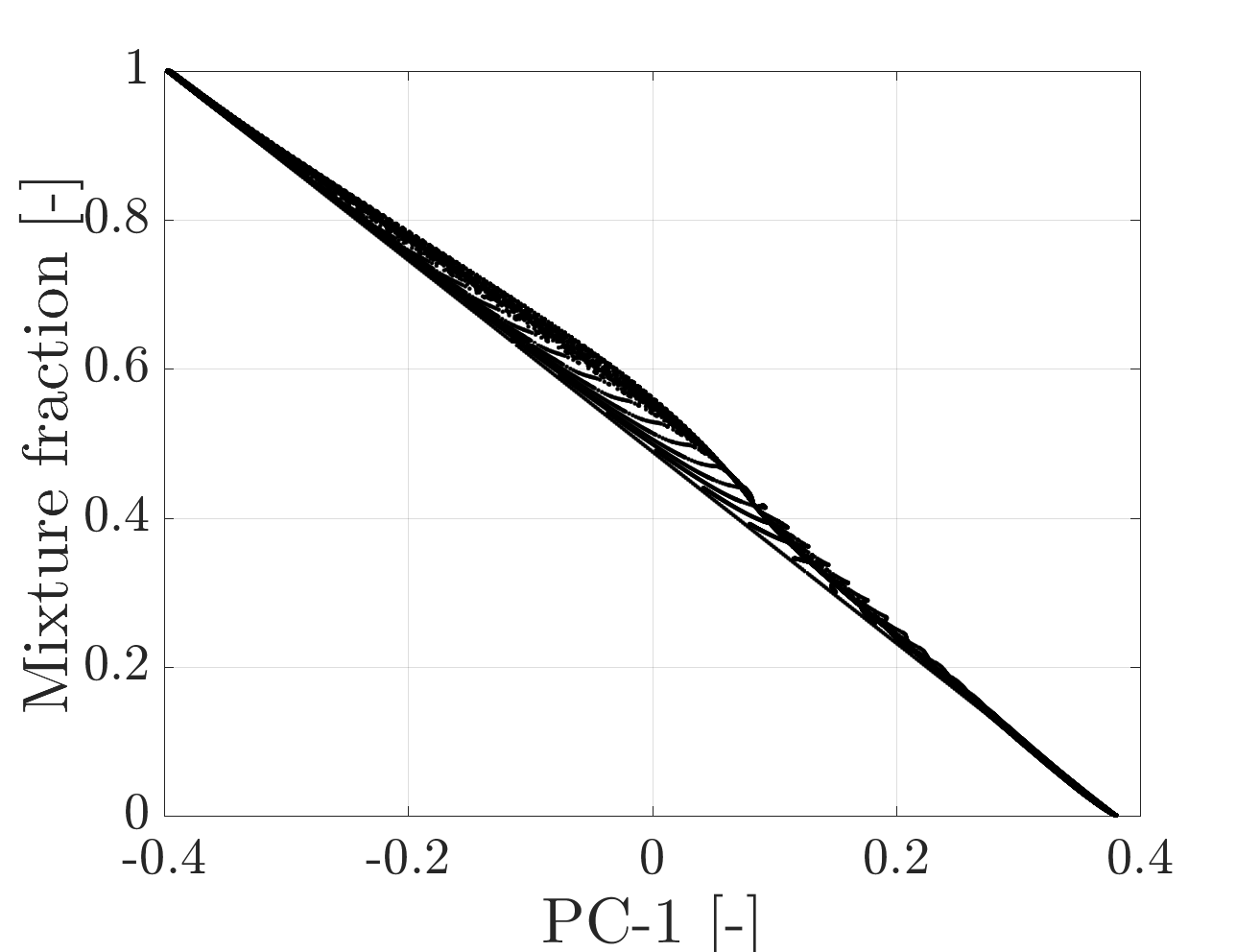}
\caption[Scatter plot of the first PC (PC-1) versus the mixture fraction variable.]
{Scatter plot of the first PC (PC-1) versus the mixture fraction variable. Reprinted from \cite{malik2020combustion} with permission from Elsevier.}
\label{fig:PC-1-F}
\end{figure}
\begin{figure}[h!]
\includegraphics[width=14pc]{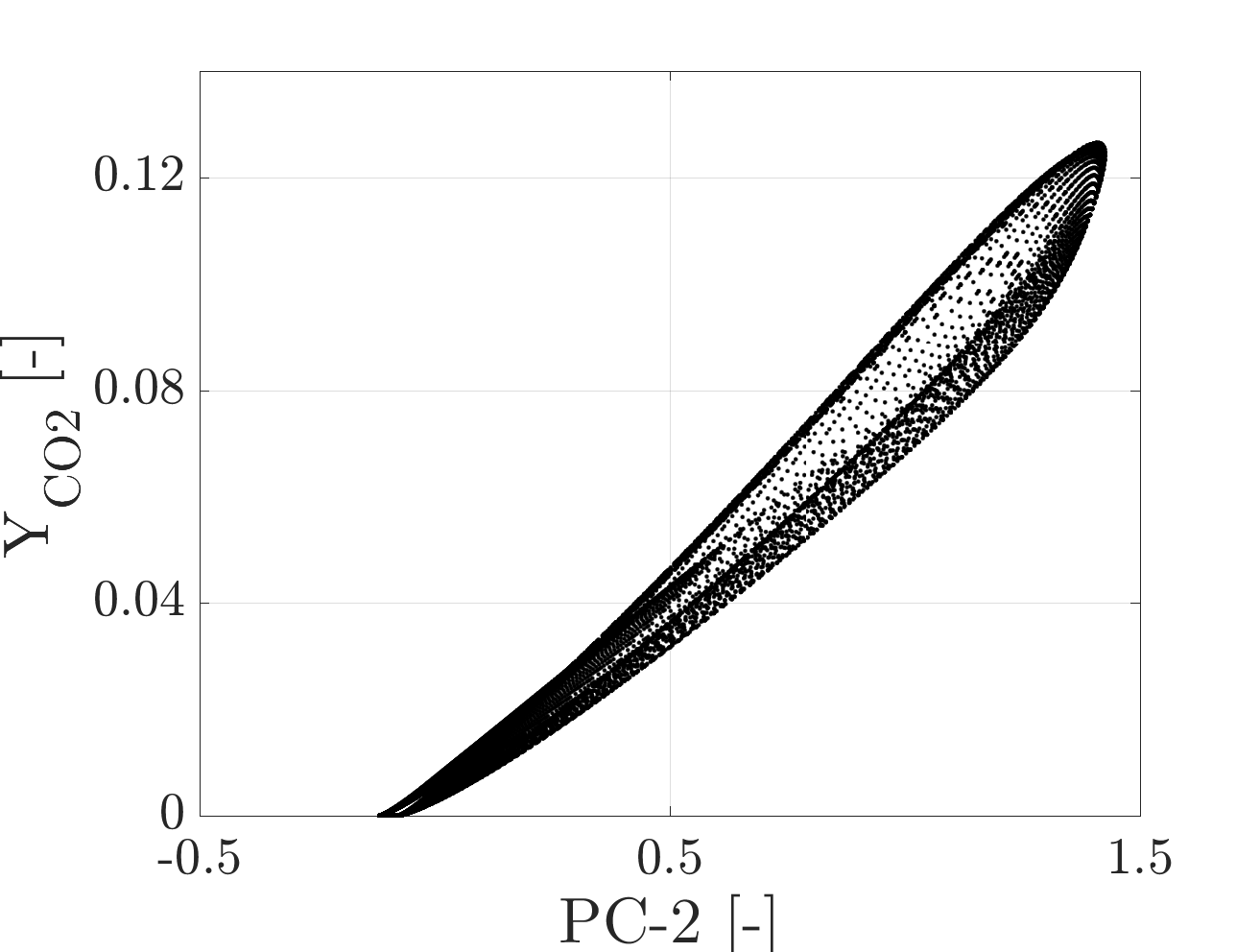}
\caption[Scatter plot of the second PC (PC-2) versus the mass fraction of $CO_2$ species, $Y_{CO_2}$.]
{Scatter plot of the second PC (PC-2) versus the mass fraction of $CO_2$ species, $Y_{CO_2}$. Reprinted from \cite{malik2020combustion} with permission from Elsevier.}
\label{fig:PC-2-YCO2}
\end{figure}
Table \ref{Table_A_5} shows the basis matrix weights obtained from the PCA analysis on the five major chemical species only. It can be seen that $\mathbf{Z}_1$ has a large positive weight for $CH_{4}$ and a large negative value for the oxidizer ($O_{2}$ and $\mathrm{N_{2}}$). This can be linked to the definition of mixture fraction from eq.(\ref{eq:mixture-fraction}). Figure \ref{fig:PC-1-F} shows a plot of the first PC highly correlated with the mixture fraction (correlation factor exceeding 99\%). Therefore, in the numerical simulation, the first PC is directly replaced by the mixture fraction, to avoid transporting a reactive scalar. The weights for $\mathbf{Z}_{2}$ also show an interesting pattern: a positive correlation for $H_{2}O$ and $CO_{2}$, and a negative correlation for $CH_{4}$, $O_{2}$, and $N_{2}$. This can be linked to a progress variable, where products have positive stoichiometric coefficients and reactants negative ones. Figure \ref{fig:PC-2-YCO2} shows a plot of the second PC highly correlated with the mass fraction of $CO_{2}$, a variable that can be attributed to the progress of reaction. It is worth pointing out that PCA identifies these patterns without any prior assumptions or knowledge of the system of interest.

\begin{figure}[h!]
\includegraphics[width=14pc]{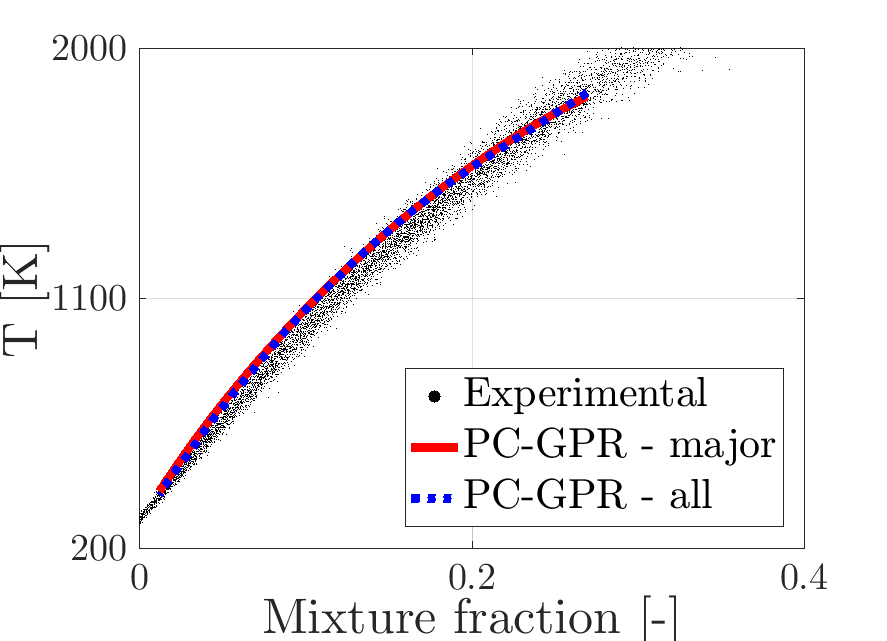}
\caption[Conditional average of the temperature at $x/D=60$.]
{Conditional average of the temperature at $x/D=60$. Reprinted from \cite{malik2020combustion} with permission from Elsevier.}
\label{fig:T-vs-MF-60}
\end{figure}

LES simulations were then performed in OpenFOAM using the PC-transport approach. The independent variables (PCs) are transported, and the state-space is recovered from the nonlinear regression using GPR. The low-Mach Navier-Stokes equations are solved on an unstructured grid, together with the PC-transport equations from eq.(\ref{eq:PC_transport}). Since the state-space was accurately regressed using $q=2$ PCs, the simulation was carried out using only $\mathbf{Z}_{1}$ and $\mathbf{Z}_{2}$ as transported PCs. More details about the numerical setting are reported in \cite{malik2020combustion}. As part of the validation of the method, we look at the temperature profiles conditioned on mixture fraction at axial location $x/D=60$ ($x=432$mm), shown in Figure \ref{fig:T-vs-MF-60}. It can be observed that the predicted temperature lies well inside the single shot experimental data points.
\begin{figure}[h!]
\includegraphics[width=14pc]{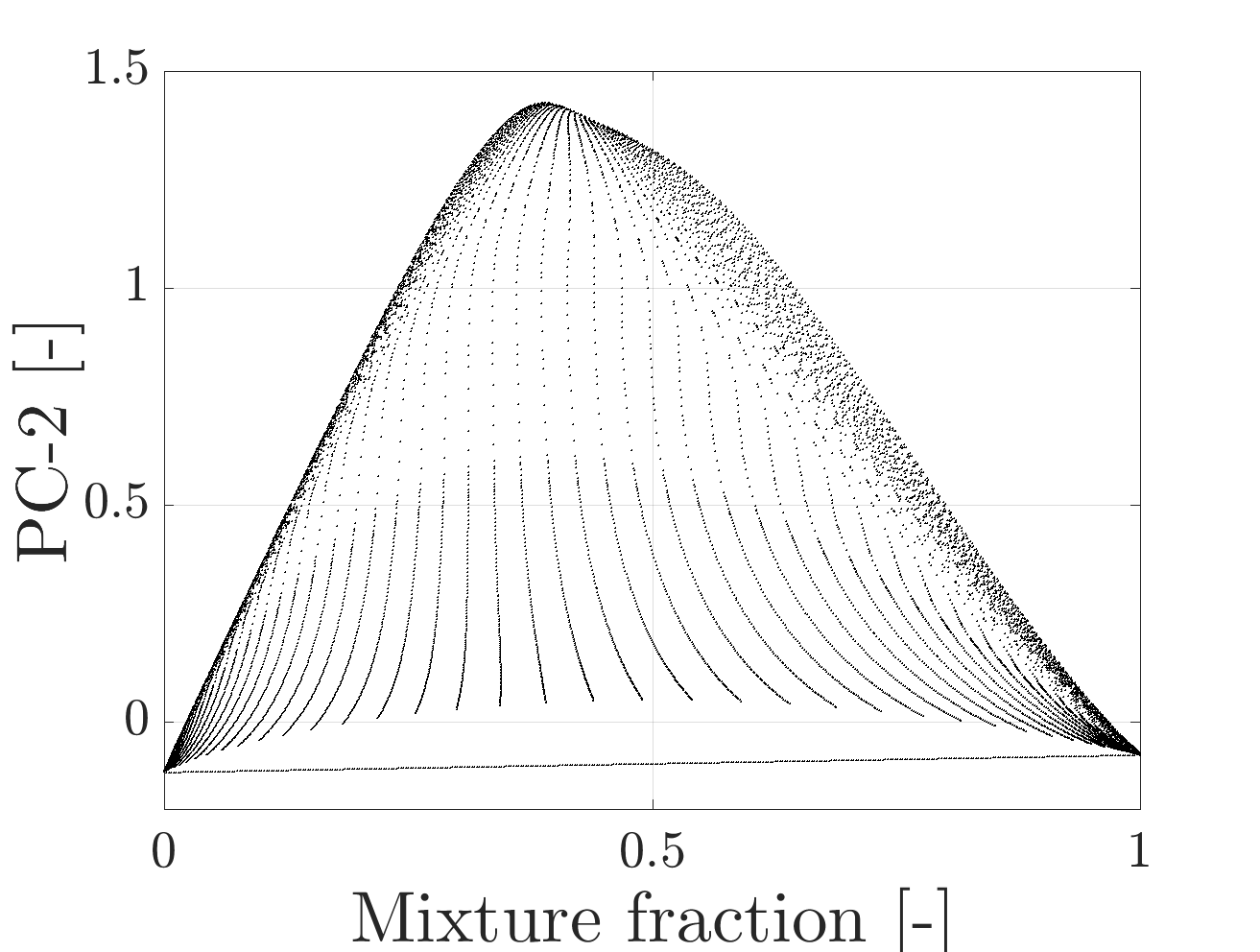}
\caption[Scatter plot of the low-dimensional combustion manifold in the space of the second PC (PC-2), and the mixture fraction variable. Manifold obtained from the training data set.]
{Scatter plot of the low-dimensional combustion manifold in the space of the second PC (PC-2), and the mixture fraction variable. Manifold obtained from the training data set. Reprinted from \cite{malik2020combustion} with permission from Elsevier.}
\label{fig:pca-manifolds-training}
\end{figure}
\begin{figure}[h!]
\includegraphics[width=14pc]{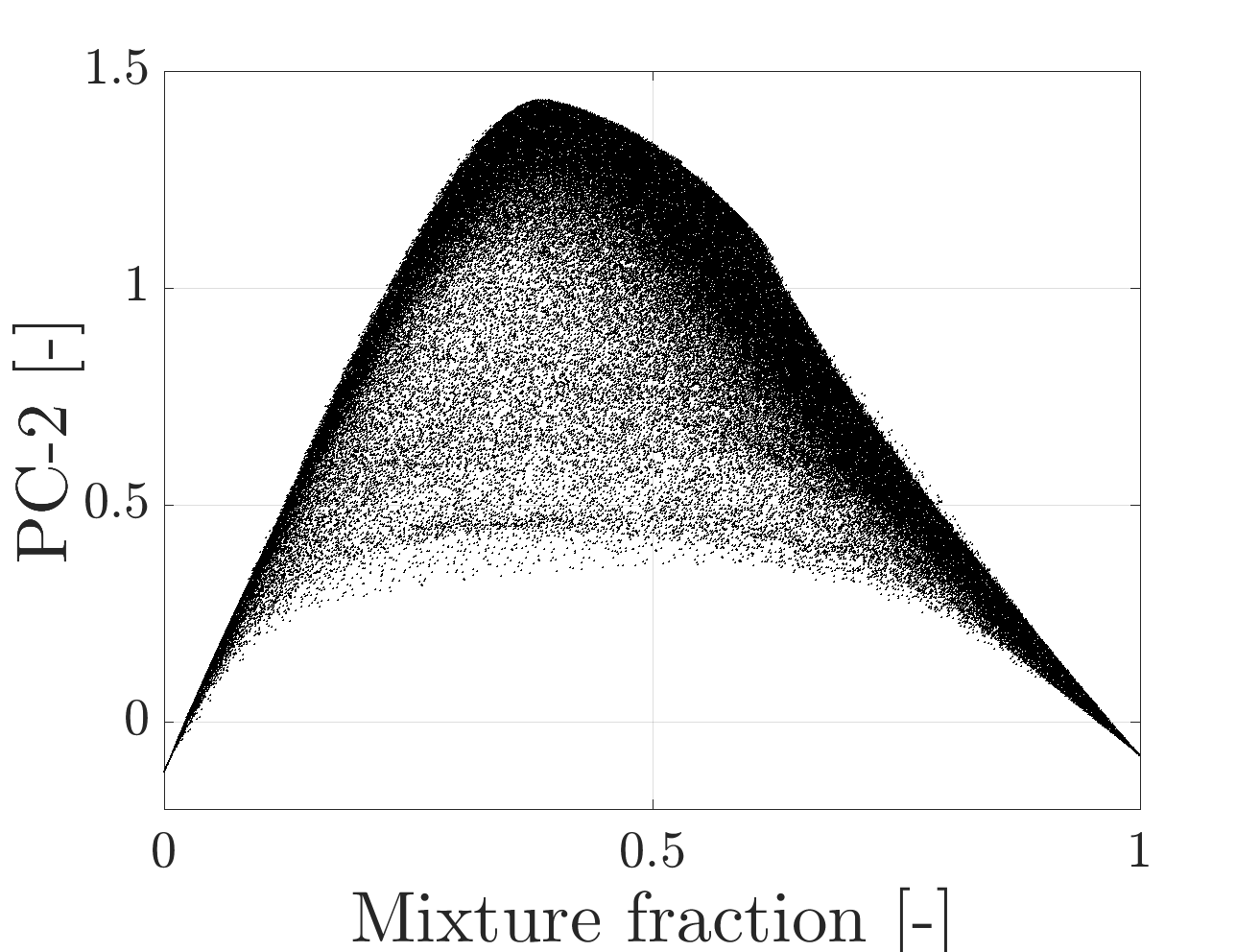}
\caption[Scatter plot of the low-dimensional combustion manifold in the space of the second PC (PC-2), and the mixture fraction variable. Manifold accessed during the simulation.]
{Scatter plot of the low-dimensional combustion manifold in the space of the second PC (PC-2), and the mixture fraction variable. Manifold accessed during the simulation. Reprinted from \cite{malik2020combustion} with permission from Elsevier.}
\label{fig:pca-manifolds-simulation}
\end{figure}

Figure \ref{fig:pca-manifolds-training} shows the original manifold obtained from the training data set, while Fig. \ref{fig:pca-manifolds-simulation} shows the manifold accessed during the simulation with major species at $t=1s$. It can be observed that the simulation does not leave the training manifold, all the points accessed are bounded inside the original training manifold. It is also apparent that most of the solution data is clustered toward the equilibrium solution, showing that the Flame D does not experience significant extinction and re-ignition.

The PC-GPR model can accurately simulate a complex multi-dimensional flame, using only 2 PCs instead of the 35 chemical species. The model can be built using inexpensive one-dimensional simulations, as long as the training data covers the potential state-space accessed during the actual simulation. The PCs stay bounded to the training manifold during the simulation, indicating that the choice of an unsteady canonical reactor ensures to span all the potential chemical states accessed during the simulation. The strength of the method resides in the fact that it does not require any prior selection of variables. Instead, it automatically extracts the most relevant variables to describe the system. From this perspective, the PC-GPR method can be regarded as a generalization of tabulated chemistry approaches \cite{Pope_small_scales}, particularly for complex systems requiring the definition of a larger number of progress variables.

\section{Chemistry Acceleration via Adaptive-Chemistry}
\label{section:sparc}

In \S\ref{section:pc_transport}, PCA was used to derive the reduced number of transport equations for the new set of variables, PCs. Regression was introduced to handle the nonlinearity of chemical source terms. In this section, we investigate the potential of local PCA (\S\ref{section:local_pca}) to classify the thermo-chemical state-space into locally homogeneous regions and apply chemical mechanism reduction locally.

\subsection{Description of the Approach}

In \cite{Ren20088165}, the authors proposed the operator-splitting strategy for application in numerical algorithms used to solve the multi-component reactive flows. With this approach, chemistry acceleration techniques can be implemented to reduce the computational cost related to the inclusion of detailed kinetic mechanisms. If we consider $\mathbf{\Psi}$ as a vector representing the temperature and mass fractions of chemical species, the transport equation for $\mathbf{\Psi}$ can be written as:
\begin{equation}\label{eq:transport_all}
    \frac{d\mathbf{\Psi}}{dt} = \mathbf{C}(\mathbf{\Psi},t) + \mathbf{D}(\mathbf{\Psi},t) +
    \mathbf{S}(\mathbf{\Psi})
\end{equation}
where $\mathbf{C}(\mathbf{\Psi},t)$ and $\mathbf{D}(\mathbf{\Psi},t)$ are the convective and diffusive transport terms, respectively, and $\mathbf{S}(\mathbf{\Psi})$ is the source term vector (representing the rates of change of $\mathbf{\Psi}$ due to chemical reactions) \cite{cuoci2013numerical}. For $N$ grid points in the simulation, we obtain a system of $N$ ordinary differential equations (ODEs).
According to the Strang splitting scheme \cite{strang1968construction}, eq.(\ref{eq:transport_all}) can be solved by grouping the contributions of convection and diffusion and integrating them separately from the chemical source term:
\begin{equation}\label{eq:operator_splitting}
\begin{cases}
\frac{d\mathbf{\Psi}^a}{dt} = \mathbf{S}(\mathbf{\Psi}^a) \\
\frac{d\mathbf{\Psi}^b}{dt} = \mathbf{C}(\mathbf{\Psi}^b{,t}) +  \mathbf{D}(\mathbf{\Psi}^b{,t})
\end{cases}
\end{equation}
where the indices $a$ and $b$ denote the separate contributions to $\mathbf{\Psi}$. In particular, three sub-steps are adopted for the numerical integration:
\begin{enumerate}
    \item \textit{Reaction step:} The ODE system corresponding to the source term $\mathbf{S}(\mathbf{\Psi}^a)$ is integrated over $\frac{\Delta t}{2}$.
    \item \textit{Transport step:} The ODE system accounting for the convection and diffusion terms $\mathbf{C}(\mathbf{\Psi}^b,t)$ and $\mathbf{D}(\mathbf{\Psi}^b,t)$ is integrated over $\Delta$t.
    \item \textit{Reaction step:} The source term $\mathbf{S}(\mathbf{\Psi}^a)$ is again integrated over $\frac{\Delta t}{2}$.
\end{enumerate}
Unlike the transport step, the two reaction steps do not require any boundary conditions and they do not have any spatial dependence. The system of $N$ ODEs from steps 1 and 3 can be solved independently from the system of $N$ ODEs from step 2. This makes the adaptive-chemistry techniques very effective and easy to implement. The idea behind the adaptive-chemistry approach is that it is possible to locally consider only a subset of the chemical species implemented in the detailed mechanism, while the remaining subset consists of the species that locally have zero concentration, or result to be not chemically active. It is thus possible to build a library of reduced mechanisms in the preprocessing step, covering the composition space, which is expected to be visited during simulation of the flame of interest \cite{Schwer2003451}, \cite{Banerjee2006619}. To build the library of reduced mechanisms, the state-space must be first partitioned into a prescribed number of clusters, and a reduced kinetic mechanism is then created separately in each cluster.
The high-dimensional space can be partitioned via the VQPCA algorithm (\S\ref{section:local_pca}), and for each cluster a reduced kinetic mechanism is generated via directed relation graph with error propagation (DRGEP) \cite{Pepiot-Desjardins200867}. At each time-step of the CFD simulation, the whole set of species in the detailed mechanism is transported. Before the chemical step, the grid points are classified \textit{on-the-fly} using the local PCA reconstruction error metrics \cite{kambhatla1997dimension}, and for each of them the appropriate reduced mechanism is selected from the library.
Although a direct relation between the CPU time and the number of species implemented in the mechanism is found for both the transport and the reaction sub-steps described in the eq.(\ref{eq:operator_splitting}), alleviating the costs related to the reaction step appears to be more important. The reaction step, in fact, results to be the most consuming part of the computation requiring up to 80 - 85\% of the CPU time \cite{cuoci2013numerical}. The full details of the overall procedure, called SPARC, as well as its validation for steady and unsteady laminar flames, can be found in \cite{d2020adaptive}.

\subsection{Application of the Approach}

\begin{figure}
\includegraphics[width=\textwidth]{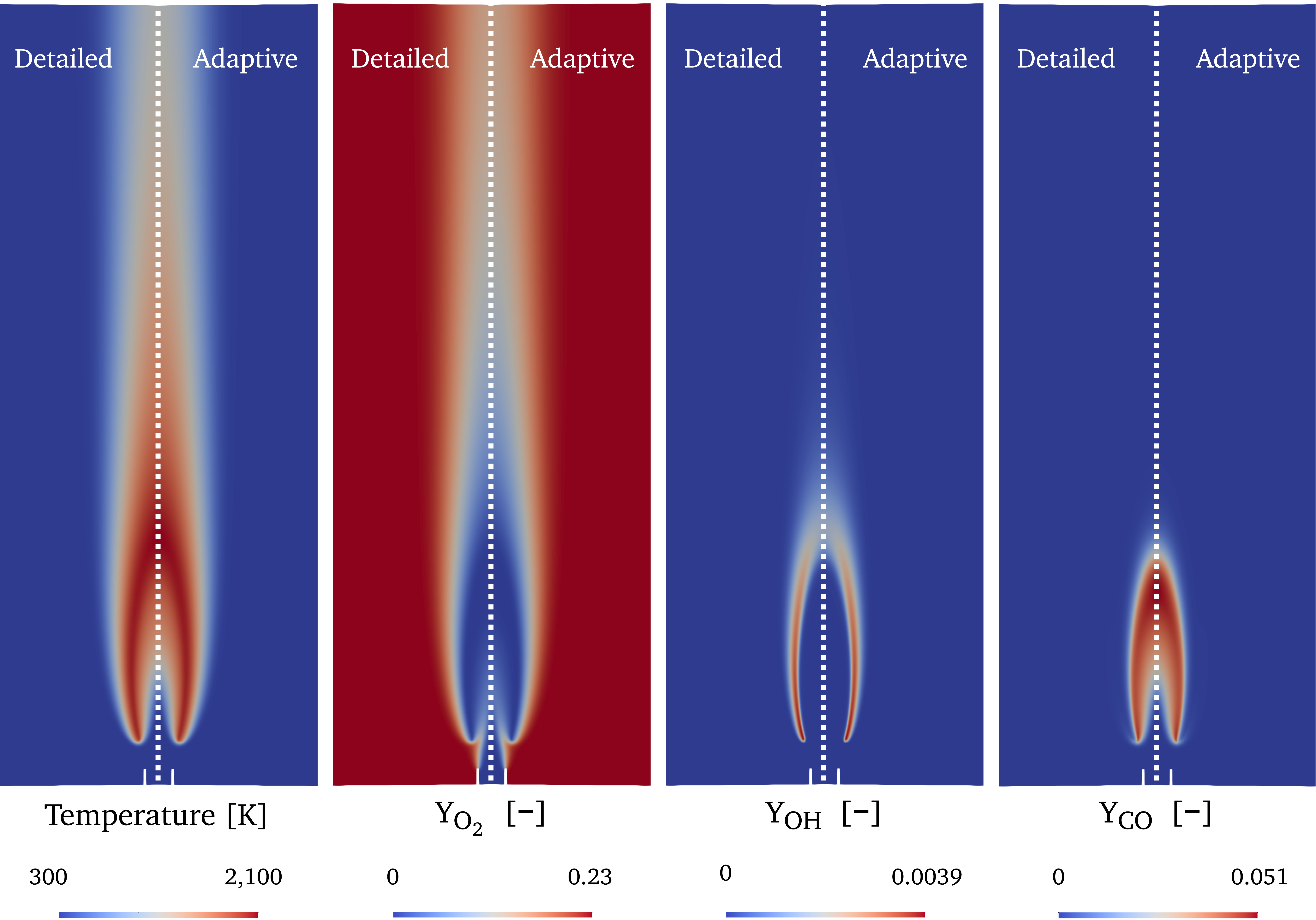}
\caption[Temperature, \ch{O2}, \ch{OH} and \ch{CO} profiles obtained from the detailed and the adaptive-chemistry simulation of an axisymmetric, non-premixed laminar nitrogen-diluted methane coflow flame.]
{Temperature, $O_2$, $OH$ and $CO$ profiles obtained from the detailed and the adaptive-chemistry simulation of an axisymmetric, non-premixed laminar nitrogen-diluted methane coflow flame.}
\label{fig:adaptive-chemistry-flames}
\end{figure}

The application of the adaptive-chemistry approach is briefly presented here for the simulation of an axisymmetric, non-premixed laminar nitrogen-diluted methane co-flow flame \cite{Mohammed1998693}.
The detailed simulation was first carried out using the \texttt{POLIMI\_C1C3\_HT\_1412} kinetic mechanism (84 species and 1698 reactions) \cite{Ranzi_2014}, and the adaptive simulations were then performed using four degrees of chemical reduction, ranging from $\epsilon_{DRGEP} = 0.03$ to $\epsilon_{DRGEP} = 0.005$. The average and the maximum number of species included in the reduced mechanisms depending on the reduction tolerance $\epsilon_{DRGEP}$ are reported in Table \ref{tab:lambda_methane}. An excellent agreement was observed for all the cases, although for $\epsilon_{DRGEP} = 0.03$ and 0.02 a negligible discrepancy was observed for the lift-off height. The inverse proportionality between the simulation accuracy and the degree of reduction, i.e., the tolerance used for DRGEP, is due to a different number of species included in the reduced mechanisms.
Figure \ref{fig:adaptive-chemistry-flames} shows a comparison between the temperature and species profiles obtained from the detailed and the adaptive simulation carried out with $\epsilon_{DRGEP} = 0.005$.
\begin{table}
\caption[The mean and maximum number of chemical species versus the uniformity coefficients for prescribed tolerances $\epsilon$ in DRGEP. The composition space was partitioned via the VQPCA algorithm.]
{The mean and maximum number of chemical species versus the uniformity coefficients for prescribed tolerances $\epsilon$ in DRGEP. The composition space was partitioned via the VQPCA algorithm.}
\label{tab:lambda_methane}
\begin{tabular}{@{}ccc@{}}
\hline
$\epsilon_{DRGEP}$ & $n_{sp}^{mean}$ & $n_{sp}^{max}$\\
\hline
0.03  & 31   & 38   \\
0.02  & 34   & 42   \\
0.01  & 39   & 44   \\
0.005 & 43    & 50   \\
\hline
\end{tabular}
\end{table}

The inclusion of a higher number of chemical species in the reduced mechanisms obviously entails an increase of the mean CPU-time required to carry out the chemical step ($\bar{\tau}_{chem}$). Nevertheless, the adaptive simulation speed-up with respect to the detailed simulation ($S_{chem}$) is large ($S_{chem} \sim 4$) even when the lowest degree of chemical reduction is adopted, as reported in Table \ref{tab:perform2dSteady}.
\begin{table}
\caption[Performances of adaptive-chemistry algorithm: average CPU-time per cell (in $ms$) for chemical step integration ($\bar{\tau}_{chem}$) and relative mean speed-up factor ($S_{chem}$) for the steady-state laminar methane flame with respect to the degree of chemistry reduction ($\epsilon_{DRGEP}$).]
{Performances of adaptive-chemistry algorithm: average CPU-time per cell (in $ms$) for chemical step integration ($\bar{\tau}_{chem}$) and relative mean speed-up factor ($S_{chem}$) for the steady-state laminar methane flame with respect to the degree of chemistry reduction ($\epsilon_{DRGEP}$).}
\label{tab:perform2dSteady}
\begin{tabular}{@{}ccc@{}}
\hline
$\epsilon_{DRGEP}$ & $\bar{\tau}_{chem}$ & $S_{chem}$\\
\hline
0.03  & 2.78   & 5.39 \\
0.02  & 3.03   & 4.94  \\
0.01  & 3.25   & 4.61 \\
0.005 & 3.78   & 3.97  \\
detailed    & 15.02  & - \\
\hline
\end{tabular}
\end{table}

If multi-dimensional simulations of the same system are not available, the SPARC model can be trained using lower-dimensional (0D or 1D) detailed simulations of the same chemical system. Reduced mechanisms can then be generated based on such training data sets and applied in the multi-dimensional (2D or 3D) adaptive simulations. In \cite{d2020adaptive}, a data set consisting of observations generated through steady-state CFDF simulations (see \S\ref{section:data-sets}), evaluated at different strain rates from $10 s^{-1}$ to $330 s^{-1}$ (the latter corresponding to extinction conditions) was also considered. This new data set consisted of $\sim 11,000$ points, corresponding to $30$ different strain rates randomly chosen in the considered interval. Also in this case, the accuracy using mechanisms obtained from a lower dimensional training data set was high, with the speed-up with respect to the detailed simulation between 4.5 and 6. More details can be found in \cite{d2020adaptive}.

\section{Available Software}

Apart from the many available MATLAB\textsuperscript{\tiny\textregistered} routines, some of which were mentioned in this chapter, two recently developed Python libraries can be used to perform data-driven ROM:

\begin{itemize}
\item \texttt{OpenMORe} \cite{DAlessio2020} is a collection of Python modules for reduced-order modeling (ROM), clustering, and classification. It incorporates many ROM techniques such as: PCA, local PCA, kernel PCA and NMF along with the varimax rotation for factor analysis. It can also be used for data clustering with VQPCA or FPCA algorithms, and it introduces utilities for evaluating the clustering solutions. The software, along with a detailed documentation and several examples, is available at \texttt{https://github.com/burn-research/OpenMORe}.
\item \texttt{PCAfold} \cite{zdybal2020pcafold} is a Python library that can be used to generate, improve and analyze low-dimensional manifolds obtained via PCA. Several novel functionalities to perform PCA on sampled data sets and analyzing the improved representation of the thermo-chemical state-space variables \cite{Zdybal2020state} were introduced. The effect of the training data preprocessing on the topology of the low-dimensional manifolds can be investigated with the novel approaches and metrics to assess the quality of manifolds \cite{Armstrong2020}, \cite{zdybal2022manifold}, \cite{zdybal2022cost}. \texttt{PCAfold} also accommodates for treatment of the thermo-chemical source terms for use in PC-transport approaches (\S\ref{section:pc_transport}). The user can find numerous illustrative examples and the associated Jupyter notebooks in the software documentation (available at \texttt{https://pcafold.readthedocs.io/}) under \textbf{Tutorials \& Demos}. The software is available at \texttt{https://gitlab.multiscale.utah.edu/common/PCAfold}.
\end{itemize}

\section{Summary}\label{section:remarks}

In this chapter, several examples of the application of data-driven techniques to multi-component reactive flow systems have been shown. Data reduction techniques, such as PCA, NMF or autoencoders also offer a way to explore hidden features of data sets (\S\ref{section:feature}). These methods are capable of detecting information in an unsupervised way and can thus aid in the process of selecting the best set of variables to effectively parameterize complex systems using fewer dimensions. Moreover, it has been shown how the aforementioned algorithms can be effectively coupled with algorithms that are widely used in the combustion community, such as DRGEP, to obtain physics-informed reduced-order models. Two applications of reduced-order modeling were presented in this chapter: reduction of the number of transport equations (\S\ref{section:pc_transport}) and reduction of large chemical mechanisms (\S\ref{section:sparc}). The main power of data-driven techniques is that the modeling can be informed by applying the technique on simple systems that are computationally cheap to obtain. Further improvement can be achieved based on the feedback coming from the validation experiments.

\section*{Acknowledgements}\label{section:ack}

Kamila Zdybał acknowledges the support of the Fonds National de la Recherche Scientifique (F.R.S.-FNRS) through the Aspirant Research Fellow grant.
Giuseppe D'Alessio acknowledges the support of the Fonds National de la Recherche Scientifique (F.R.S.-FNRS) through the FRIA fellowship.
Gianmarco Aversano, Mohammad Rafi Malik and Alessandro Parente acknowledge the funding from the European Research Council (ERC) under the European Union’s Horizon 2020 research and innovation programme under grant agreement No 714605.

\end{document}